\relax
\documentclass[letterpaper]{article}
\usepackage{aaai17}
\usepackage{times}
\usepackage{helvet}
\usepackage{courier}
\usepackage{amsfonts}
\usepackage{amsmath}
\usepackage{url}
\usepackage{graphicx}
\newtheorem{mydef}{Definition}
\newtheorem{mythm}{Theorem}
\begin{document}
%
\title{Exploring and measuring non-linear correlations:\\ Copulas, Lightspeed Transportation and Clustering}
\author{Gautier Marti\\
Hellebore Capital Ltd\\
Ecole Polytechnique\\
\And
S\'ebastien Andler\\
ENS de Lyon\\
Hellebore Capital Ltd
\And
Frank Nielsen\\
Ecole Polytechnique\\
LIX - UMR 7161
\And
Philippe Donnat\\
Hellebore Capital Ltd\\
Michelin House, London
}
\maketitle
\begin{abstract}
We propose a methodology to explore and measure the pairwise correlations 
that exist between variables in a dataset.
The methodology leverages copulas for encoding dependence between two variables, 
state-of-the-art optimal transport for providing a relevant geometry to the copulas, 
and clustering for summarizing the main dependence patterns found between the variables.
Some of the clusters centers can be used to parameterize a novel dependence coefficient 
which can target or forget specific dependence patterns.
Finally, we illustrate and benchmark the methodology on several datasets. 
Code and numerical experiments are available online for reproducible research. 
\end{abstract}

\section{Introduction}


Pearson's correlation coefficient which estimates linear dependence between two variables is still the mainstream tool
for measuring variable correlations in science and engineering.
However, its shortcomings are well-documented in the statistics literature: 
not robust to outliers; 
not invariant to monotone transformations of the variables;
can take value 0 whereas variables are strongly dependent;
only relevant when variables are jointly normally distributed.
A large but under-exploited literature in statistics and machine learning 
has expanded recently to alleviate these issues \cite{reshef2011detecting,szekely2009brownian,sejdinovic2013equivalence}.
An underlying idea to many of the dependence coefficients is to compute a distance $D(P(X,Y),P(X)P(Y))$ 
between the joint distribution $P(X,Y)$ of variables $X$, $Y$ 
and $P(X)P(Y)$ the product of marginal distributions encoding the independence.
For example, choosing $D = \mathrm{KL}$ (Kullback-Leibler divergence), we end up with the Mutual Information (MI) measure, well-known in information theory.
Thus, one can detect all the dependences between $X$ and $Y$ since the distance will be greater than 0 as soon as
$P(X,Y)$ is different from $P(X)P(Y)$.
Then, the dependence literature focus has shifted toward the new concept of ``equitability'' \cite{kinney2014equitability}: 
How can one quantify the strength of a statistical association between two variables without bias for relationships of a specific form?
Many researchers now aim at designing and proving that their proposed measures are indeed equitable \cite{reshef2013equitability,ding2013copula,chang2016robust}.
This is \textit{not} what we look for in this article. 
But, on the contrary, we want to target specific dependence patterns and ignore others.
We want to target dependence which are relevant to such or such problem, 
and forget about the dependence which are not in the scope of the problems at hand, 
or even worse which may be spurious associations (pure chance or artifacts in the data).
The latter will be detected with an equitable dependence measure since they are deviation from independence,
and will be given as much weight as the interesting ones.
Rather than using the biases for specific dependence of several coefficients, 
we propose a dependence coefficient that can be parameterized by a set of {\em target-dependences}, and a set of {\em forget-dependences}.
Sets of target and forget dependences can be built using expert hypotheses, 
or by leveraging the centers of clusters resulting from an exploratory clustering of the pairwise dependences.
To achieve this goal, we will leverage three tools: 
copulas, 
optimal transportation,
and clustering.
Whereas clustering, the task of grouping a set of objects in such a way that objects in the same group (also called cluster)
are more similar to each other than those in different groups, is common knowledge in the machine learning community,
copulas and optimal transportation are not yet mainstream tools.
Copulas have recently gained attention in machine learning \cite{elidan2013copulas},
and several copula-based dependence measures have been proposed for improving 
feature selection methods \cite{ghahramani2012copula,lopez2013randomized,chang2016robust}.
Optimal transport may be more familiar to computer scientists working in computer vision 
since it is the underlying theory of the Earth Mover's Distance \cite{rubner2000earth}.
Until very recently, optimal transportation distances between distributions were not deemed relevant for 
machine learning applications since the best computational cost known was super-cubic 
to the number of bins used for discretizing the distribution supports
which grows itself exponentially with the dimension.
A mere distance evaluation could take several seconds!
In this article, we leverage recent computational breakthroughs detailed in \cite{cuturi2013sinkhorn} which
make their use practical in machine learning.

\section{Background on Copulas and Optimal Transport}

\subsection{Copulas}

Copulas are functions that couple multivariate distribution functions to their one-dimensional marginal distribution functions \cite{nelsen2013introduction}.
In this article, we will only consider bivariate copulas, but most of the results and the methodology presented hold in the multivariate setting,
at the cost of a much higher computational burden which is for now a bit unrealistic.

\begin{mythm}[Sklar's Theorem \cite{sklar1959fonctions}]
For any random vector $X = (X_i,X_j)$ having continuous marginal cumulative distribution functions $F_i, F_j$ respectively, 
its joint cumulative distribution $F$ is uniquely expressed as $F(X_i,X_j) = C(F_i(X_i),F_j(X_j))$, where $C$, 
the bivariate distribution of uniform marginals $U_i,U_j := F_i(X_i),F_j(X_j)$, is known as the copula of $X$.
\end{mythm}

\begin{figure}
\begin{center}
\includegraphics[width=0.16\textwidth]{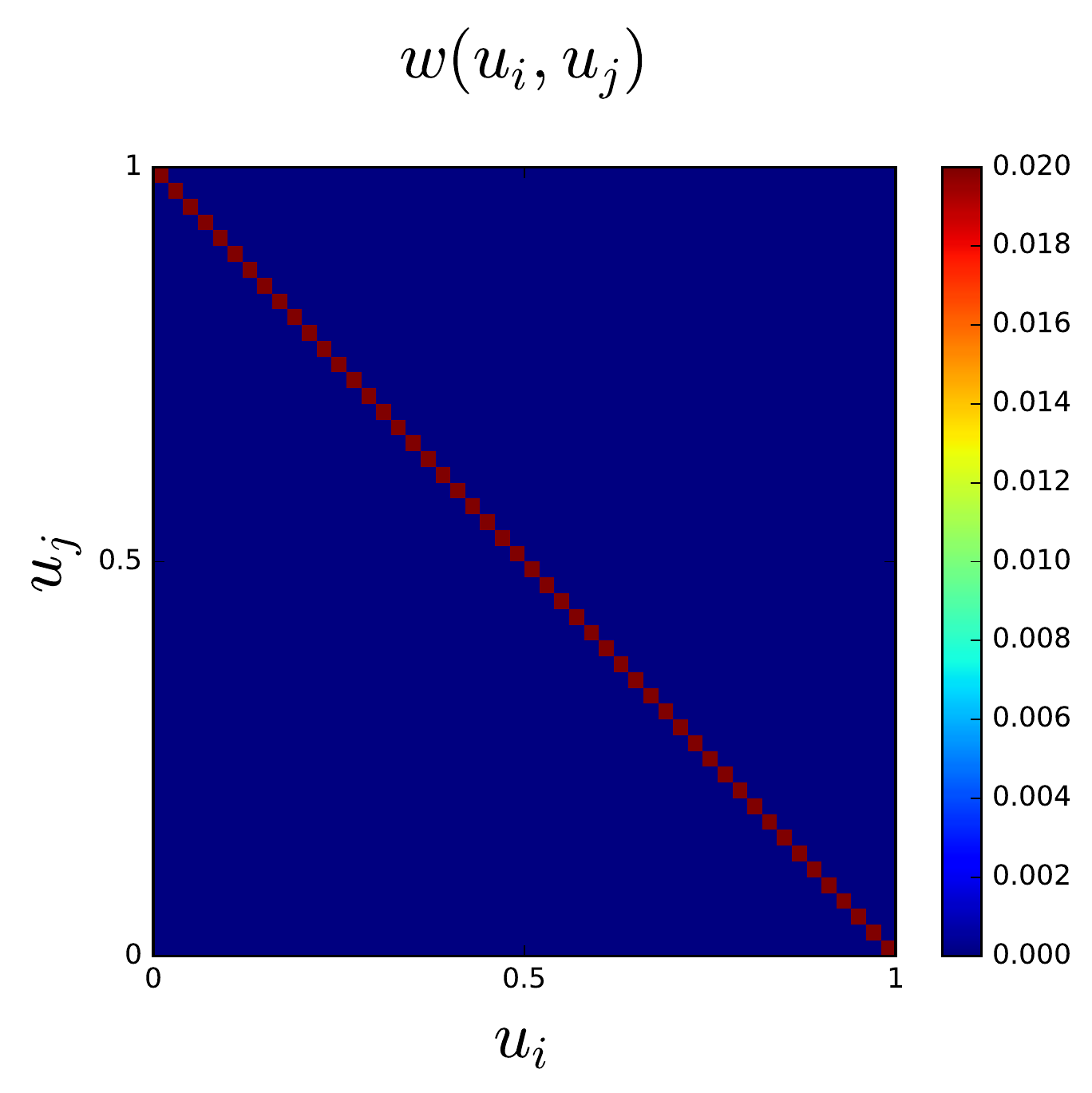}
\includegraphics[width=0.16\textwidth]{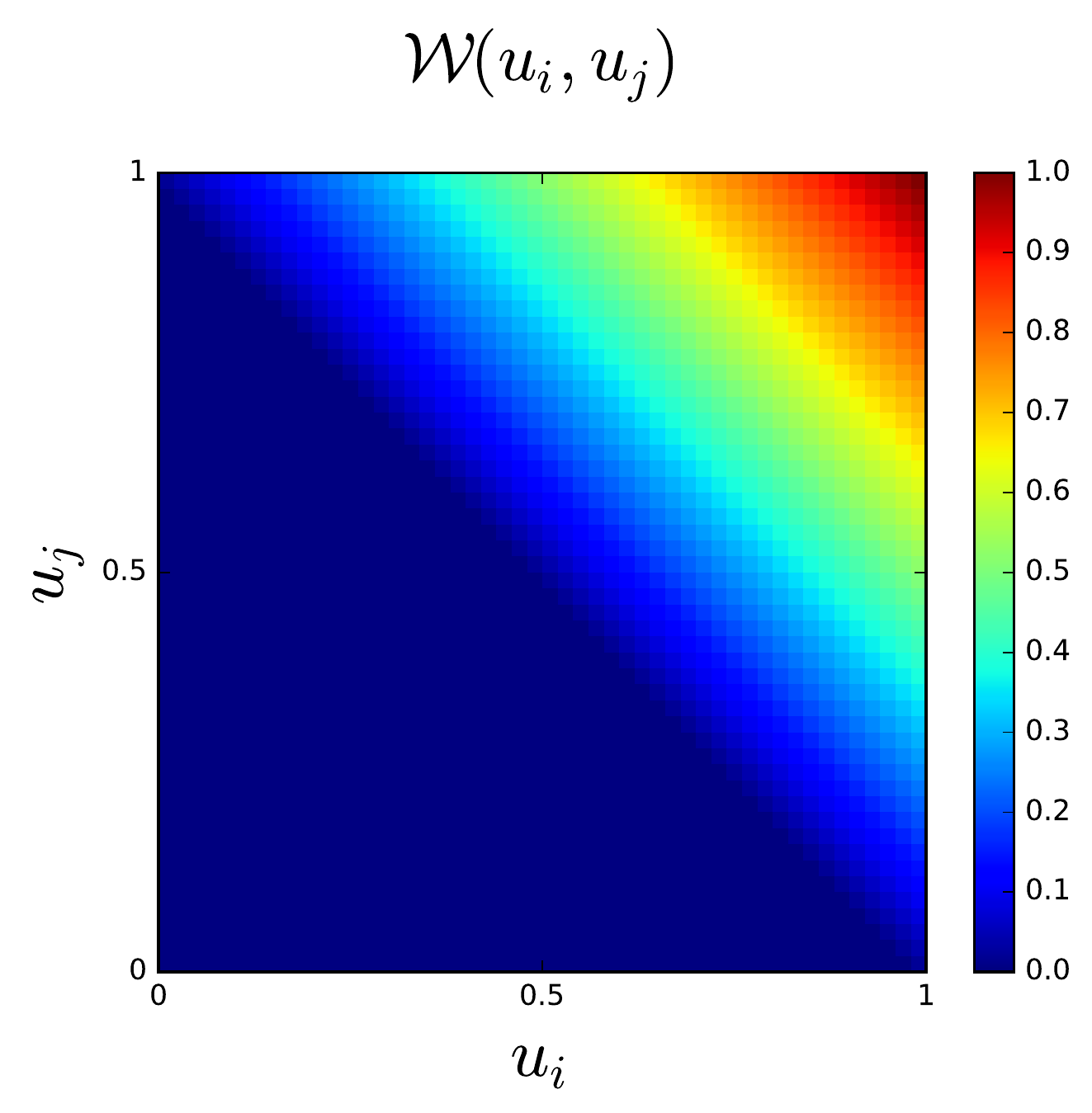}
\includegraphics[width=0.16\textwidth]{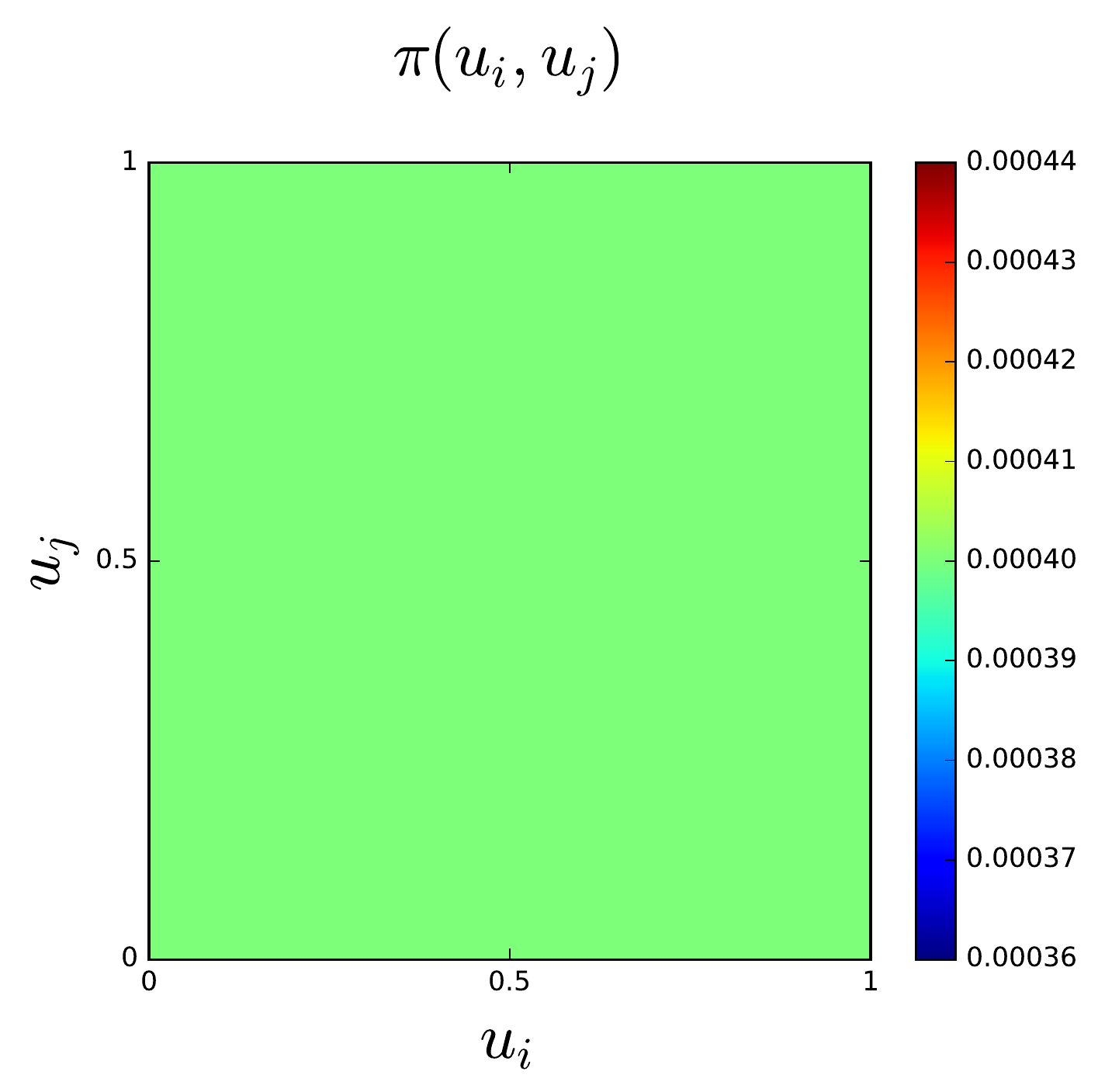}
\includegraphics[width=0.16\textwidth]{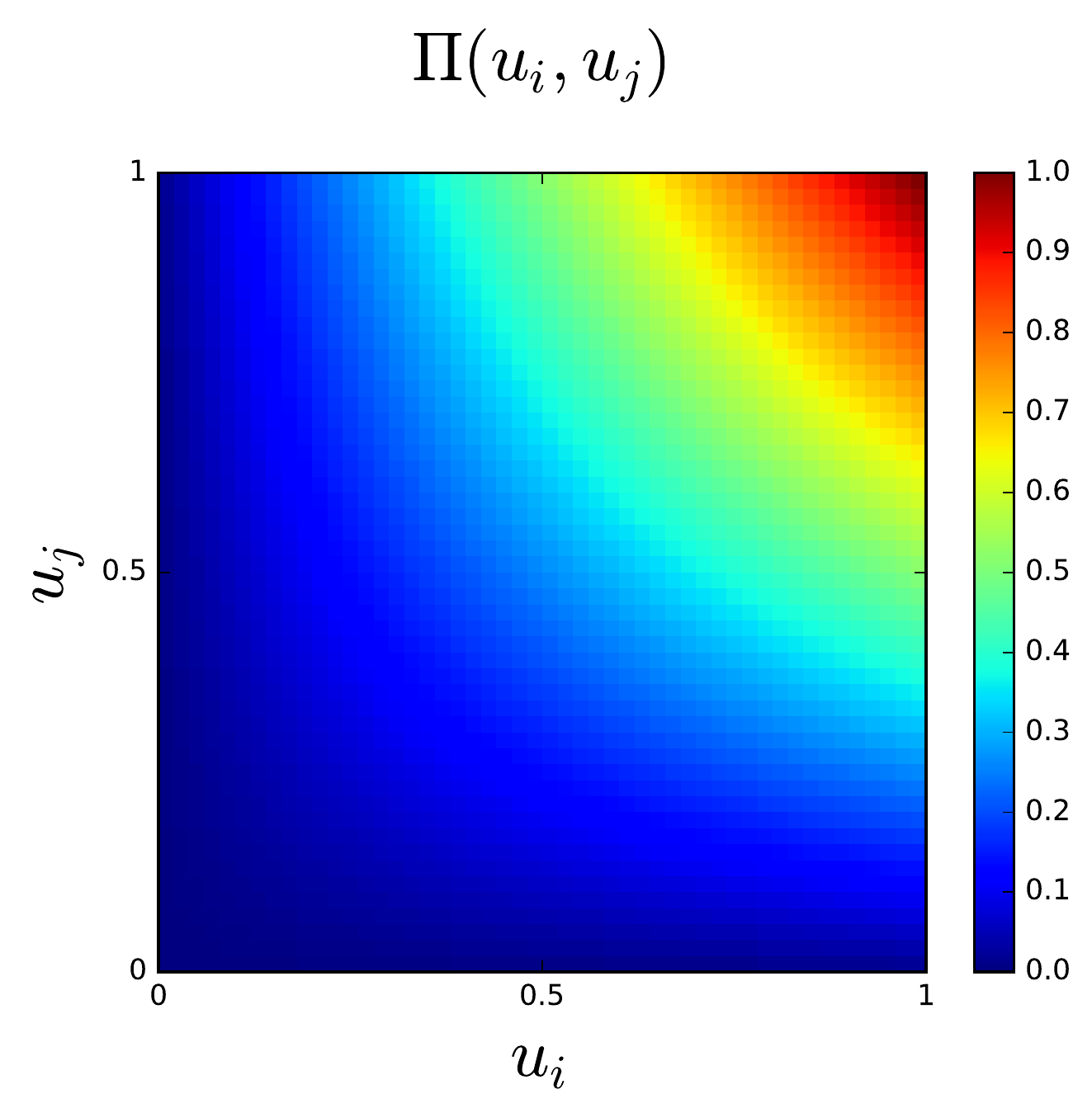}
\includegraphics[width=0.16\textwidth]{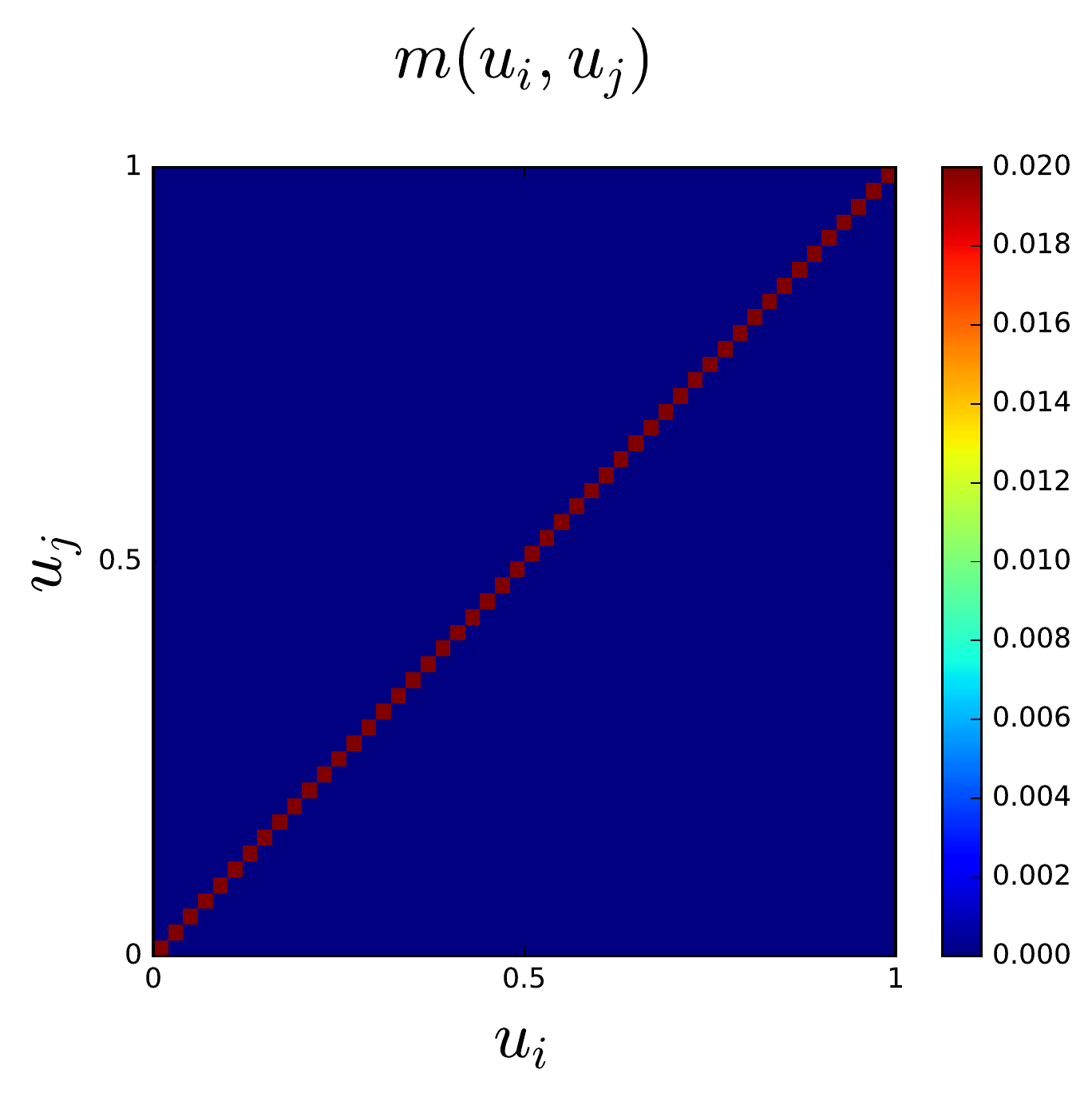}
\includegraphics[width=0.16\textwidth]{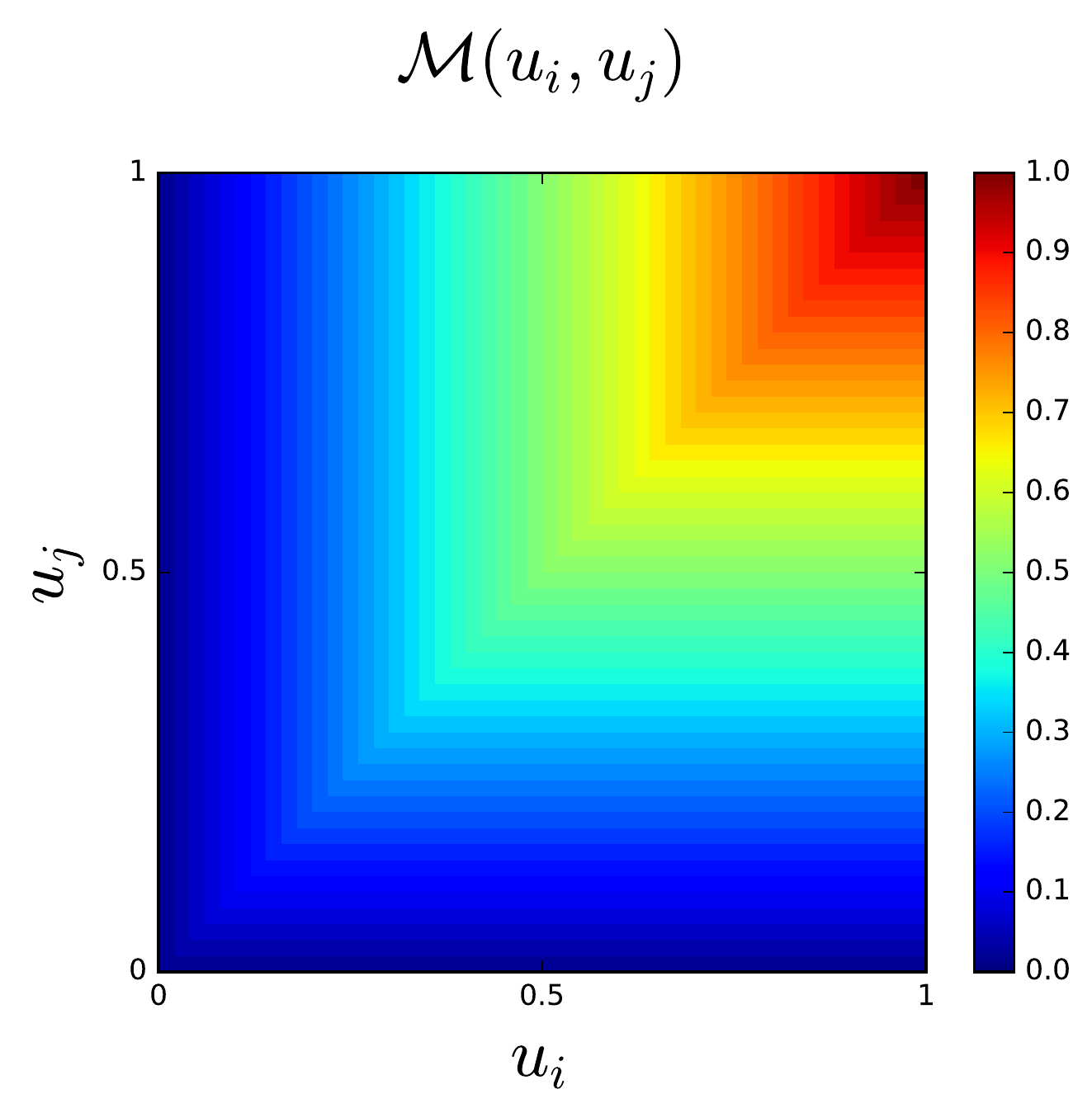}
\end{center}
\caption{Copulas measure (left column) and cumulative distribution function (right column) heatmaps for negative dependence (first row), independence (second row), i.e. the uniform distribution over $[0,1]^2$, and positive dependence (third row)}
\label{fig:copula_bounds}
\end{figure}

Copulas are central for studying the dependence between random variables: their uniform marginals jointly encode all the dependence.
They allow to study scale-free measures of dependence and are \textit{invariant to monotonous transformations of the variables}.
Some copulas play a major role in the measure of dependence, namely $\mathcal{W}$ and $\mathcal{M}$ the Fr\'echet-Hoeffding copula bounds,
and the independence copula $\Pi(u_i,u_j) = u_i u_j$ (depicted in Figure~\ref{fig:copula_bounds}).

\begin{mydef}[Fr\'echet-Hoeffding copula bounds]
For any copula $C : [0,1]^2 \rightarrow [0,1]$ and any $(u_i,u_j) \in [0,1]^2$ the following bounds hold:
\begin{equation}\label{eq:bounds}
\mathcal{W}(u_i,u_j) \leq C(u_i,u_j) \leq \mathcal{M}(u_i,u_j),
\end{equation}
where $\mathcal{W}(u_i,u_j) = \max \left\{ u_i + u_j - 1, 0 \right\}$ is the copula for countermonotonic random variables
and $\mathcal{M}(u_i,u_j) = \min \left\{ u_i, u_j \right\}$ is the copula for comonotonic random variables.
\end{mydef}

Many correlation coefficients can actually be expressed as a distance between the data copula and one of these reference copulas.
For example, the Spearman (rank) correlation $\rho_S$ which is usually understood as $\rho_S(X_i,X_j) = \rho(F_i(X_i),F_j(X_j))$, i.e. 
the linear dependence of the probability integral transformed variables (rank-transformed data), 
can also be viewed as an average distance between the copula $C$ of $(X_i,X_j)$ 
and the independence copula $\Pi$: 
$\rho_S(X_i,X_j) = 12\int\int_{[0,1]^2} \left(C(u_i,u_j) - u_iu_j \right)du_idu_j$ \cite{nelsen2013introduction}. 
Moreover, since $|u_i-u_j|/\sqrt{2}$ is the distance between point $(u_i,u_j)$
to the diagonal (the measure of the positive dependence copula), 
one can rewrite $\rho_S(X_i,X_j) = 12\int\int_{[0,1]^2} \left(C(u_i,u_j) - u_iu_j \right)du_idu_j = 
12 \int\int_{[0,1]^2} u_iu_j dC(u_i,u_j) - 3 = 1 - 6\int\int_{[0,1]^2} (u_i-u_j)^2 dC(u_i,u_j)$ \cite{liebscher2014copula}. 
Thus, Spearman correlation can also be viewed as measuring a deviation from the monotonically increasing dependence 
to the data copula using a quadratic distance.
\textit{We will leverage this idea to propose our dependence-parameterized dependence coefficient.}

Notice that when working with empirical data, we do not know a priori the margins $F_i$ for applying the probability integral transform $U_i := F_i(X_i)$.
Deheuvels in \cite{deheuvels1979fonction} has introduced a practical estimator for the uniform margins and the underlying copula, the empirical copula transform.

\begin{mydef}[Empirical Copula Transform]

Let $(X_i^t,X_j^t)$, $t = 1,\ldots,T$, be $T$ observations from a random vector $(X_i,X_j)$ with continuous margins.
Since one cannot directly obtain the corresponding copula observations $(U_i^t,U_j^t) := (F_i(X_i^t),F_j(X_j^t))$, where $t = 1,\ldots,T$, without knowing a priori $F_i$, one can instead estimate the empirical margins
$F_i^T(x) = \frac{1}{T} \sum_{t=1}^T \textbf{1}(X_i^t \leq x)$, 
to obtain the $T$ empirical observations
$(\tilde{U_i^t},\tilde{U_j^t}) := (F_i^T(X_i^t),F_j^T(X_j^t))$. Equivalently, since $\tilde{U_i^t} = R_i^t / T$, $R_i^t$ being the rank of observation $X_i^t$, the empirical copula transform can be considered as the normalized rank transform.
\end{mydef}

Notice that the empirical copula transform is fast to compute, sorting arrays of length $T$ can be done in $O(T \log T)$,
consistent and converges fast to the underlying copula \cite{deheuvels1981asymptotic}, \cite{ghahramani2012copula}.

As motivated in the introduction, we want to compare and summarize 
the pairwise empirical dependence structure (empirical bivariate copulas) of many variables.
This brings the following questions: 
How can we compare two such copulas? 
What is a relevant representative of a set of empirical copulas?
Which geometries are relevant for clustering these empirical distributions, and which are not?

\subsection{Optimal Transport}

In \cite{Mart1606:Optimal}, authors illustrate in a parametric setting using Gaussian copulas that
common divergences (such as Kullback-Leibler, Jeffreys, Hellinger, Bhattacharyya) are not relevant 
for clustering these distributions, especially when dependence is high. 
These information divergences are only defined for absolutely continuous measures whereas some copulas
have no density (e.g. the one for positive dependence).
In practice, when working with frequency histograms, it gets worse: 
One has to pre-process the empirical measures with a kernel density estimator before computing these divergences.
On the contrary, optimal transport distances are well-defined for both discrete (e.g. empirical) and continuous measures.

The idea of optimal transport is intuitive.
It was first formulated by Gaspard Monge in 1781 \cite{monge1781memoire} as a problem to efficiently level the ground: Given that work is measured by the distance multiplied by the amount of dirt displaced, what is the minimum amount of work required to level the ground? 
Optimal transport plans and distances give the answer to this problem.


In practice, empirical distributions can be represented by histograms. We follow notations from \cite{cuturi2013sinkhorn}.
Let $r$, $c$ be two histograms in the probability simplex $\Sigma_m = \{x \in \mathbb{R}_{+}^{m}~:~x^\top 1_m = 1\}$.
Let $U(r,c) = \{P \in \mathbb{R}_{+}^{m \times m}~|~P1_m = r, P^\top 1_m = c\}$ be the transportation polytope of $r$ and $c$, 
that is the set containing all possible transport plans between $r$ and $c$.

\begin{mydef}[Optimal Transport]
Given a $m \times m$ cost matrix $M$, the cost of mapping $r$ to $c$ using a transportation matrix $P$ can be quantified as
$\langle P, M \rangle_F$, where $\langle \cdot, \cdot \rangle_F$ is the Frobenius dot-product.
The optimal transport between $r$ and $c$ given transportation cost $M$ is thus:
\begin{equation}\label{eq:ot_def}
d_M(r,c) := \min_{P \in U(r,c)} \langle P, M\rangle_F.
\end{equation}
\end{mydef}
Whenever $M$ belongs to the cone of distance matrices, the optimum of the transportation problem $d_M(r,c)$ is itself a distance.


\textbf{Lightspeed transportation.} Optimal transport distances suffer from a computational burden scaling in $O(m^3 \log m)$ which has prevented 
their widespread use in machine learning: A mere distance computation between two high-dimensional histograms can take several seconds.
In \cite{cuturi2013sinkhorn}, Cuturi provides a solution to this problem: He restrains the polytope $U(r,c)$ of all possible transport plans
between $r$ and $c$ to a Kullback-Leibler ball $U_{\alpha}(r,c) \subset U(r,c)$, where
$U_{\alpha}(r,c) = \{P \in U(r,c)~|~\mathrm{KL}(P\|rc^\top) \leq \alpha\}.$
He then shows that it amounts to perform an entropic regularization (recently generalized to many more regularizers in \cite{muzellec2016tsallis,dessein2016regularized}) of the optimal transportation problem
whose solution is smoother and less deterministic.
The regularized optimal transportation problem is now strictly convex,
and can be solved efficiently using the Sinkhorn-Knopp iterative algorithm
which exhibits linear convergence.
Its solution is the Sinkhorn distance \cite{cuturi2013sinkhorn}:
\begin{equation}\label{eq:reg_ot_def}
d_{M,\alpha}(r,c) := \min_{P \in U_{\alpha}(r,c)} \langle P, M \rangle_F,
\end{equation}
and its dual $d_{M}^\lambda(r,c)$: $\forall \alpha > 0, \exists \lambda > 0,$
\begin{equation}\label{eq:dual_reg_ot_def}
d_{M,\alpha}(r,c) = d_{M}^\lambda(r,c) := \langle P^\lambda, M \rangle_F,
\end{equation}
where $P^\lambda = \mathrm{argmin}_{P \in U(r,c)} \langle P, M \rangle_F - \frac{1}{\lambda}h(P)$, and $h$ is the entropy function.

In the following, we will leverage the dual-Sinkhorn distances for comparing,
clustering and computing the clusters centers \cite{DBLP:conf/icml/CuturiD14} of a set of copulas at full speed.

\section{A methodology to explore and measure non-linear correlations}

\begin{figure}
\begin{center}
\includegraphics[width=\linewidth]{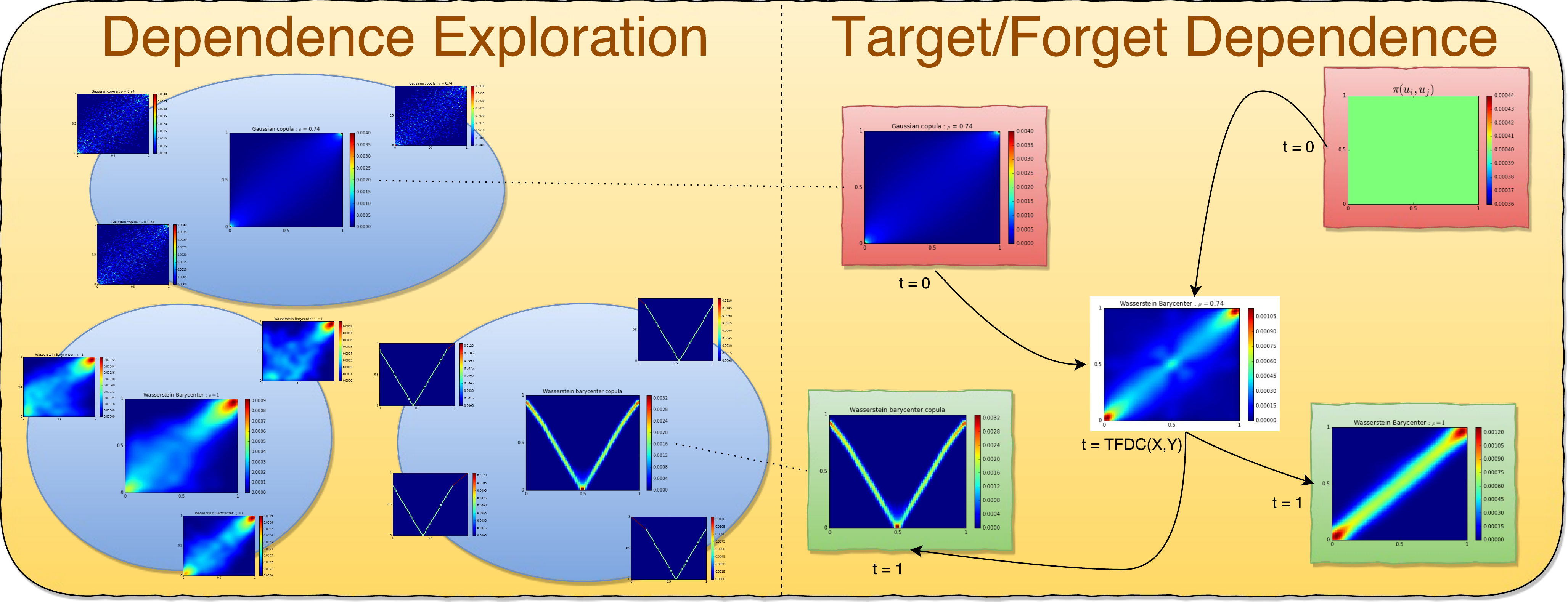}
\caption{Exploration (left panel) and measure (right panel) of non-linear correlations. Exploration consists in finding clusters of similar copulas, visualizing their centroids, and eventually using them to assess the dependence of given variables represented by their copula}
\label{fig:methodo}
\end{center}
\end{figure}

We propose an approach to explore and measure non-linear correlations between $N$ variables $X_1,\ldots,X_N$ in a dataset.
These $N$ variables can be, for instance, time series or features.
The methodology presented (which is summarized in Figure~\ref{fig:methodo}) is twofold, and consists of:
(i) an exploratory part of the pairwise dependence between variables,
(ii) the parameterization and use of a novel dependence coefficient.




\subsection{Using transportation of copulas as a measure of correlations}

In this section, we leverage and extend the idea presented in our short introduction to copulas: 
correlation coefficients can be viewed as a distance between the data-copula and the
Fr\'echet-Hoeffding bounds or the independence copula. The distance involved is usually an $\ell_p$ Minkowski metric distance.
In the following, we will:
\begin{itemize}
\item replace the $\ell_p$ distance by an optimal transport distance between measures,
\item parameterize a dependence coefficient with other copulas than the Fr\'echet-Hoeffding bounds or the independence one.
\end{itemize}

Using the optimal transport distance between copulas, we now propose a dependence coefficient 
which is parameterized by two sets of copulas: \textit{target} copulas and \textit{forget} copulas.

\begin{mydef}[Target/Forget Dependence Coefficient]
Let $\{C_l^-\}_l$ be the set of forget-dependence copulas. 
Let $\{C_k^+\}_k$ be the set of target-dependence copulas.
Let $C$ be the copula of $(X_i,X_j)$. Let $d_M$ be an optimal transport distance parameterized by a ground metric $M$.
We define the Target/Forget Dependence Coefficient as:\\
$\mathrm{TFDC}\left(X_i,X_j; \{C_k^+\}_k, \{C_l^-\}_l\right) :=$
\begin{equation}\label{eq:tfdc_def}
\frac{\min_l d_M(C_l^-,C)}{\min_l d_M(C_l^-,C) + \min_k d_M(C, C_k^+)} \in [0,1].
\end{equation}
\end{mydef}

Using this definition, we obtain: $\mathrm{TFDC}\left(X_i,X_j; \{C_k^+\}_k, \{C_l^-\}_l\right) = 0 \Leftrightarrow C \in \{C_l^-\}_l$, 
$\mathrm{TFDC}\left(X_i,X_j; \{C_k^+\}_k, \{C_l^-\}_l\right) = 1 \Leftrightarrow C \in \{C_k^+\}_k$.

\begin{figure}
\begin{center}
\includegraphics[width=0.6\linewidth]{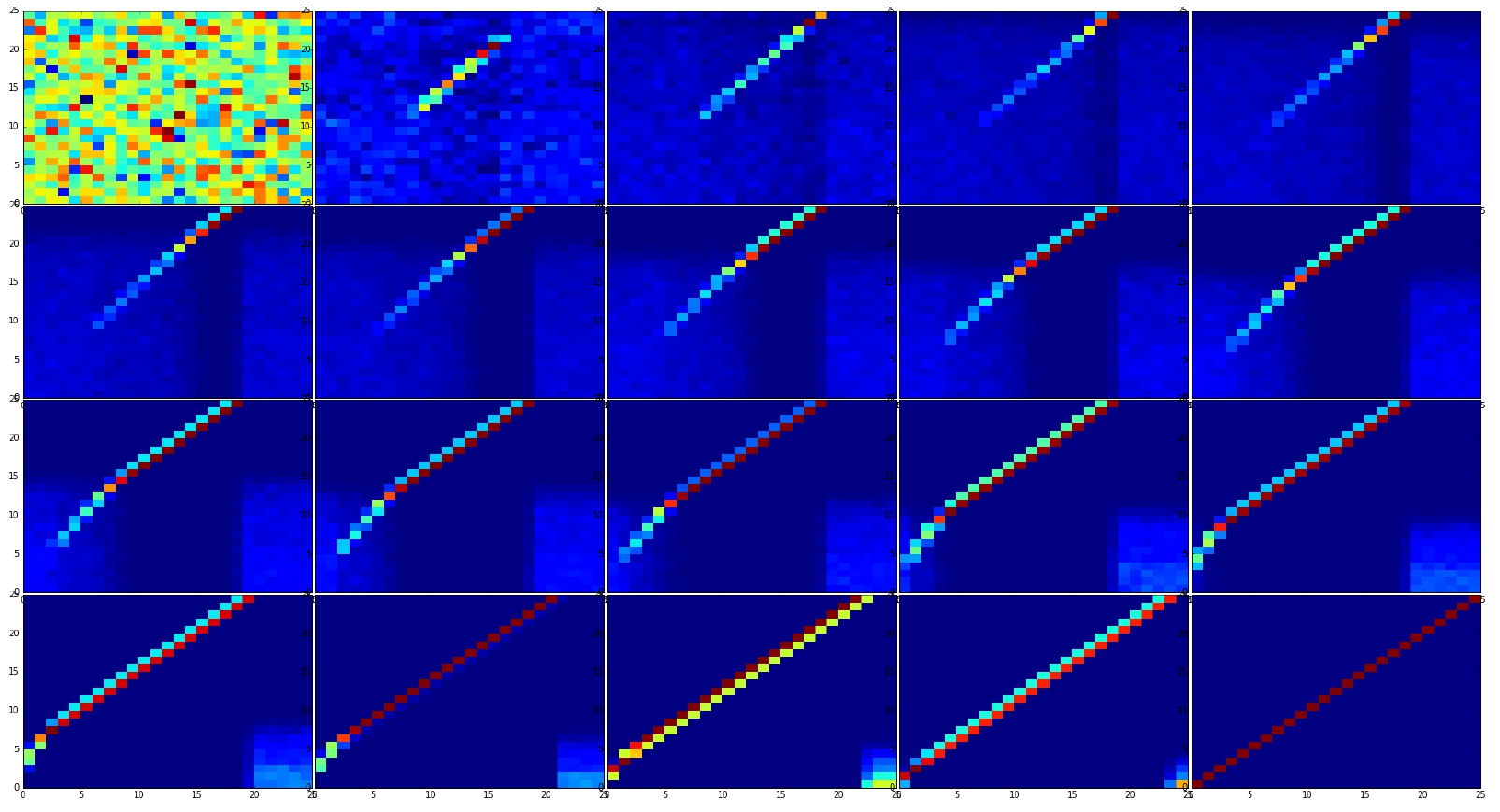}
\includegraphics[width=0.38\linewidth]{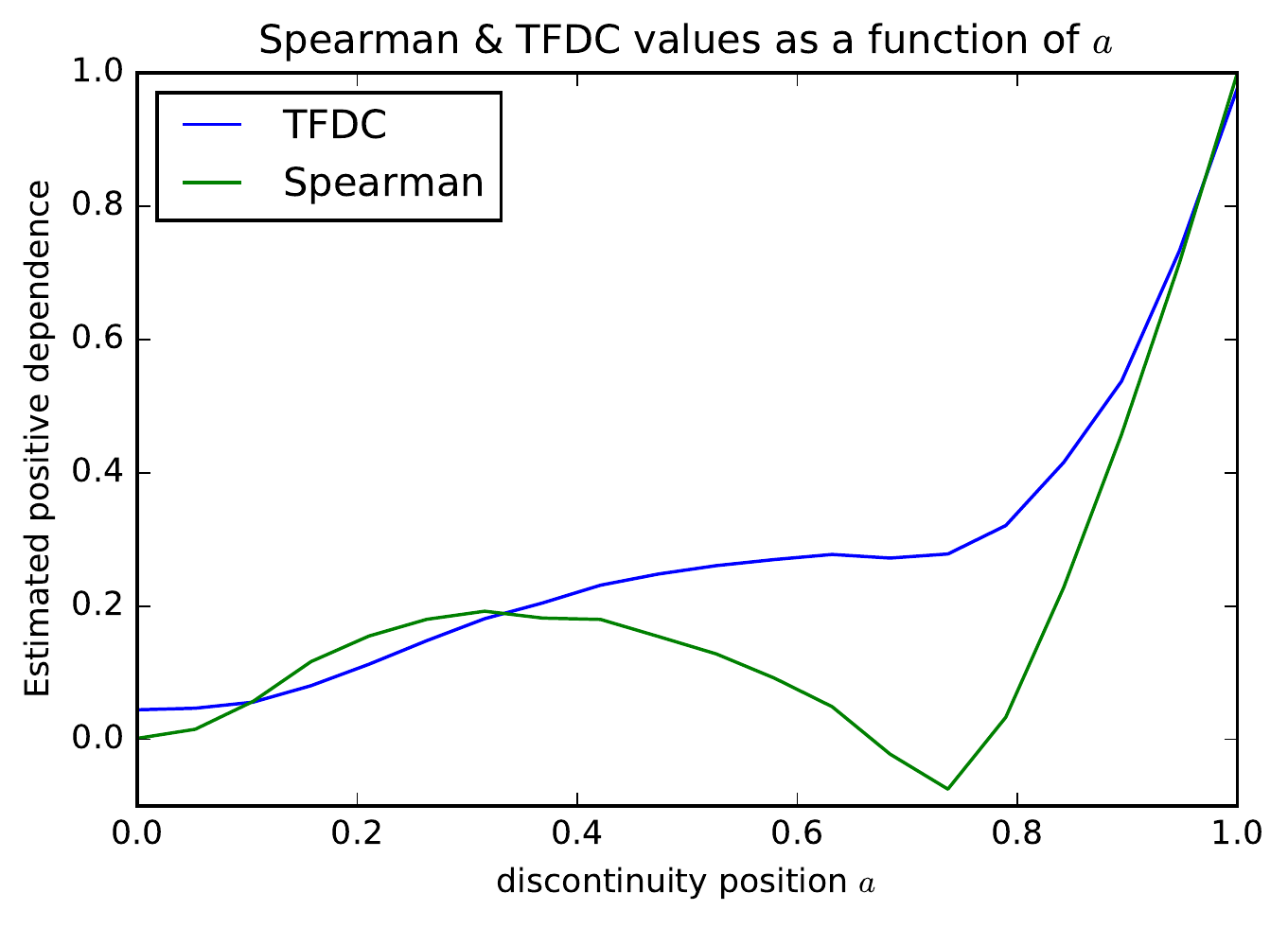}
\end{center}
\caption{Empirical copulas for $(X,Y)$ where $X=Z\mathbf{1}_{Z<a}+\epsilon_X\mathbf{1}_{Z>a}$, $Y=Z\mathbf{1}_{Z<a+0.25}+\epsilon_Y\mathbf{1}_{Z>a+0.25}$, $a = 0,0.05,\ldots,0.95,1$, 
and where $Z$ is uniform on $[0,1]$ and $\epsilon_X, \epsilon_Y$ are independent noises (left figure). Top left is an empirical copula for independence ($a = 0$), bottom right is the copula for perfect positive dependence ($a = 1$). Parameter $a$ is increasing from top to bottom, and from left to right;
TFDC and Spearman coefficients estimated between $X$ and $Y$ as a function of $a$ (right figure). For $a = 0.75$, Spearman coefficient yields a negative value, yet $X=Y$ over $[0,a]$}
\label{fig:discont_exp}
\end{figure}

\textbf{Example.} A standard correlation coefficient can be obtained by setting the forget-dependence set to the independence copula, and the target-dependence set to the Fr\'echet-Hoeffding bounds.
How does it compare to the Spearman correlation? 
In Figure~\ref{fig:discont_exp}, we display how the two coefficients behave
on a simple numerical experiment:
$X=Z\mathbf{1}_{Z<a}+\epsilon_X\mathbf{1}_{Z>a}$, $Y=Z\mathbf{1}_{Z<a+0.25}+\epsilon_Y\mathbf{1}_{Z>a+0.25}$, 
where $Z$ is uniform on $[0,1]$ and $\epsilon_X, \epsilon_Y$ are independent noises. That is $X = Y$ over $[0,a]$. Notice that for $a = 0.75$, Spearman coefficient takes a negative value.
We may thus prefer the monotonically increasing behaviour of the TFDC 
to the Spearman one.


\subsection{How to choose, design and build targets?}

We now propose two alternatives for choosing, designing and building the \textit{target} and \textit{forget} copulas:
an exploratory data-driven approach and an hypotheses testing approach.

\subsubsection{Data-driven: Clustering of copulas}

\begin{figure}
\begin{center}
\includegraphics[width=0.242\linewidth]{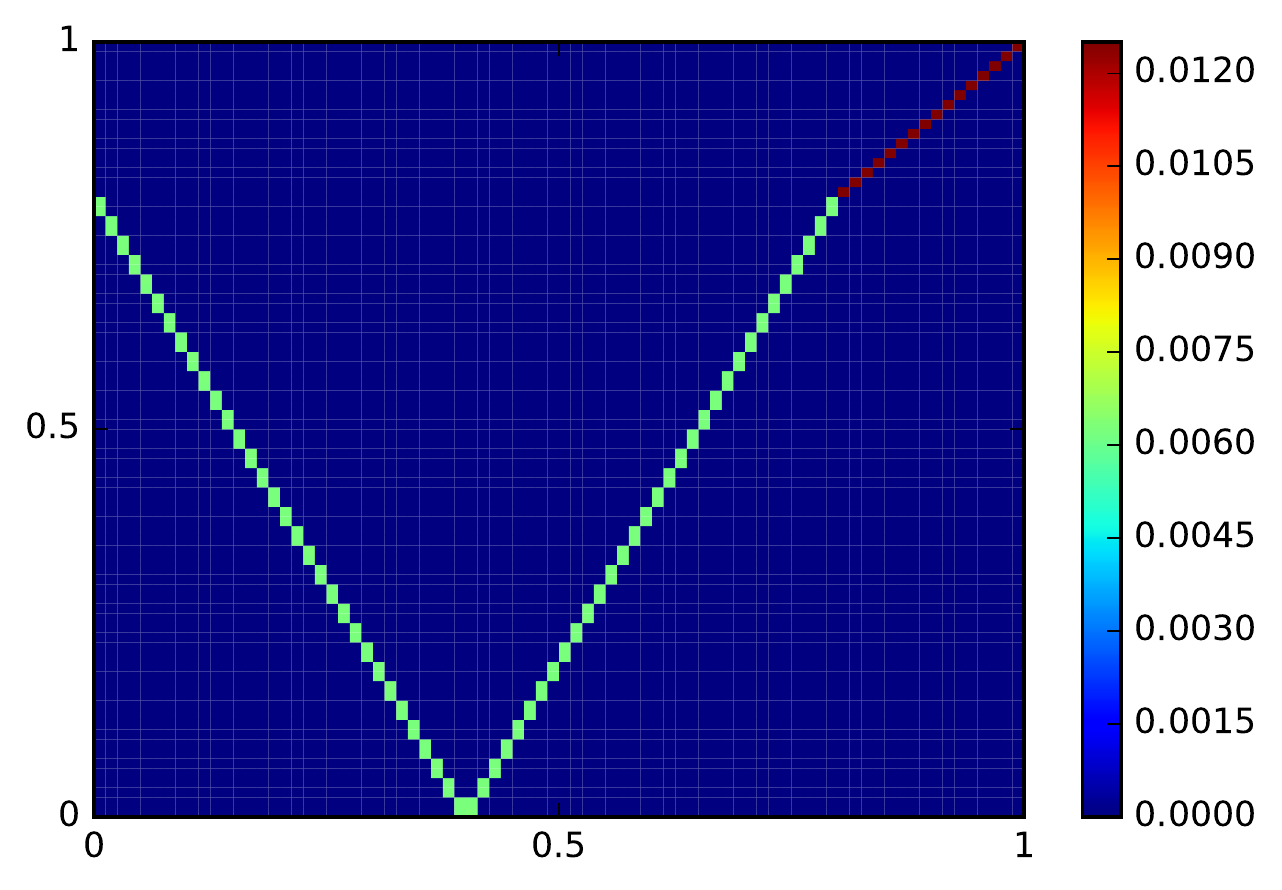}
\includegraphics[width=0.242\linewidth]{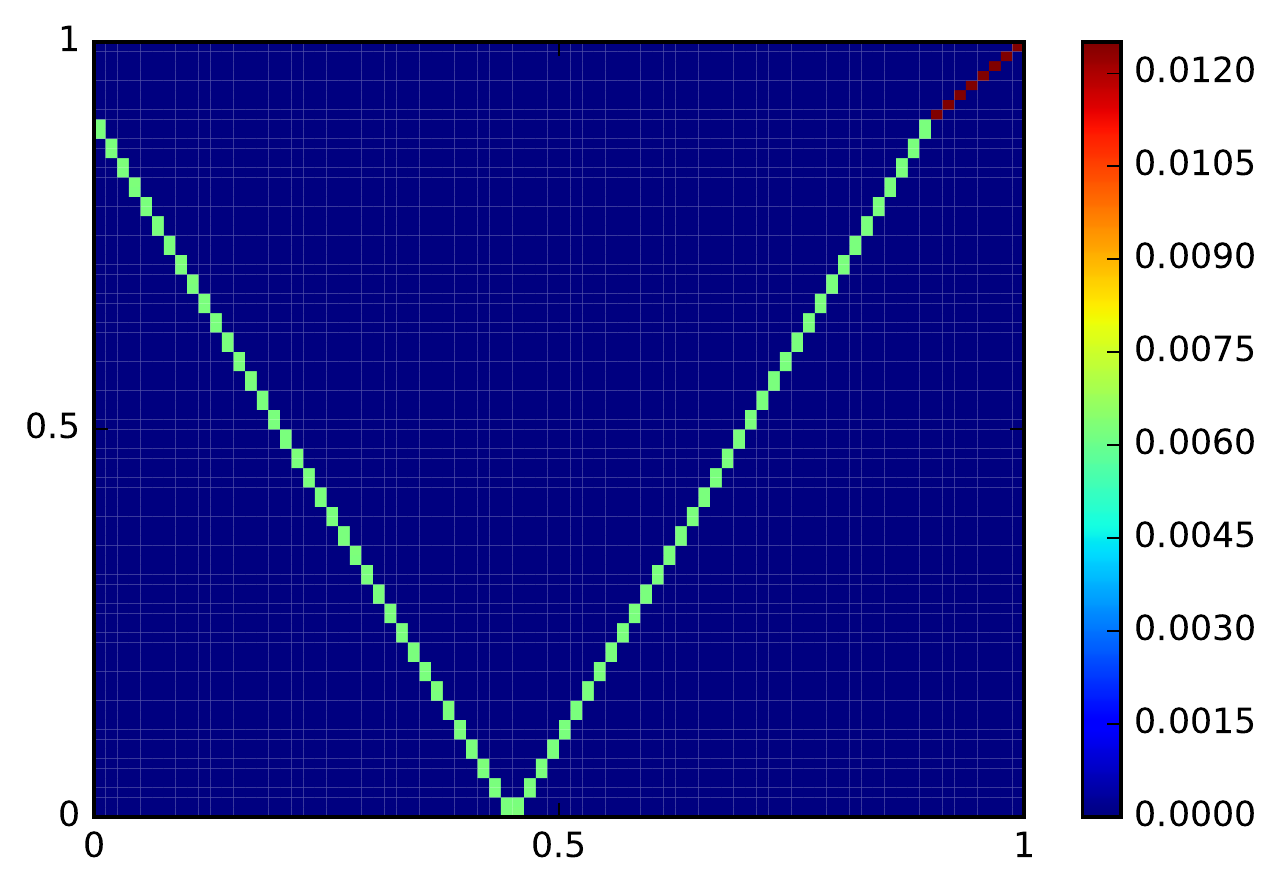}
\includegraphics[width=0.242\linewidth]{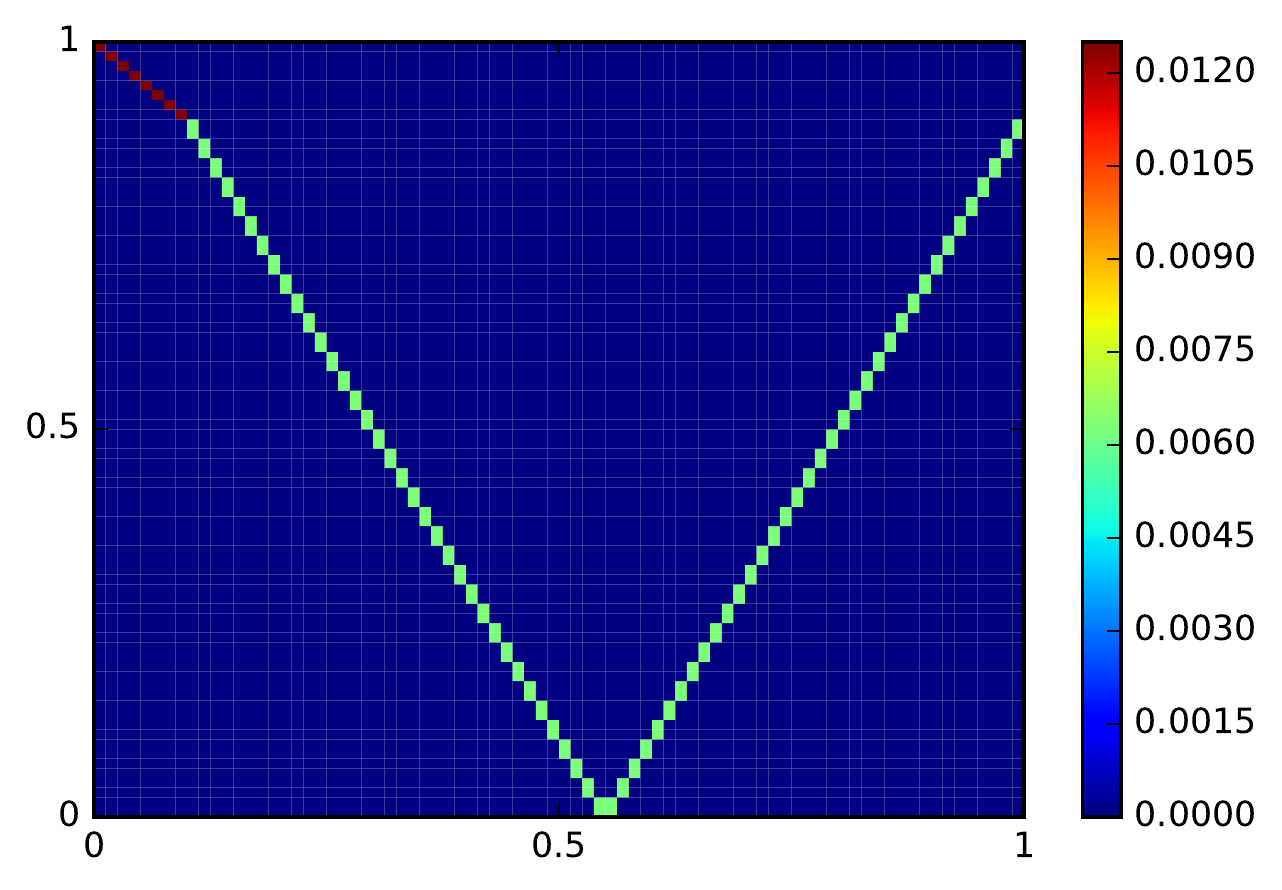}
\includegraphics[width=0.242\linewidth]{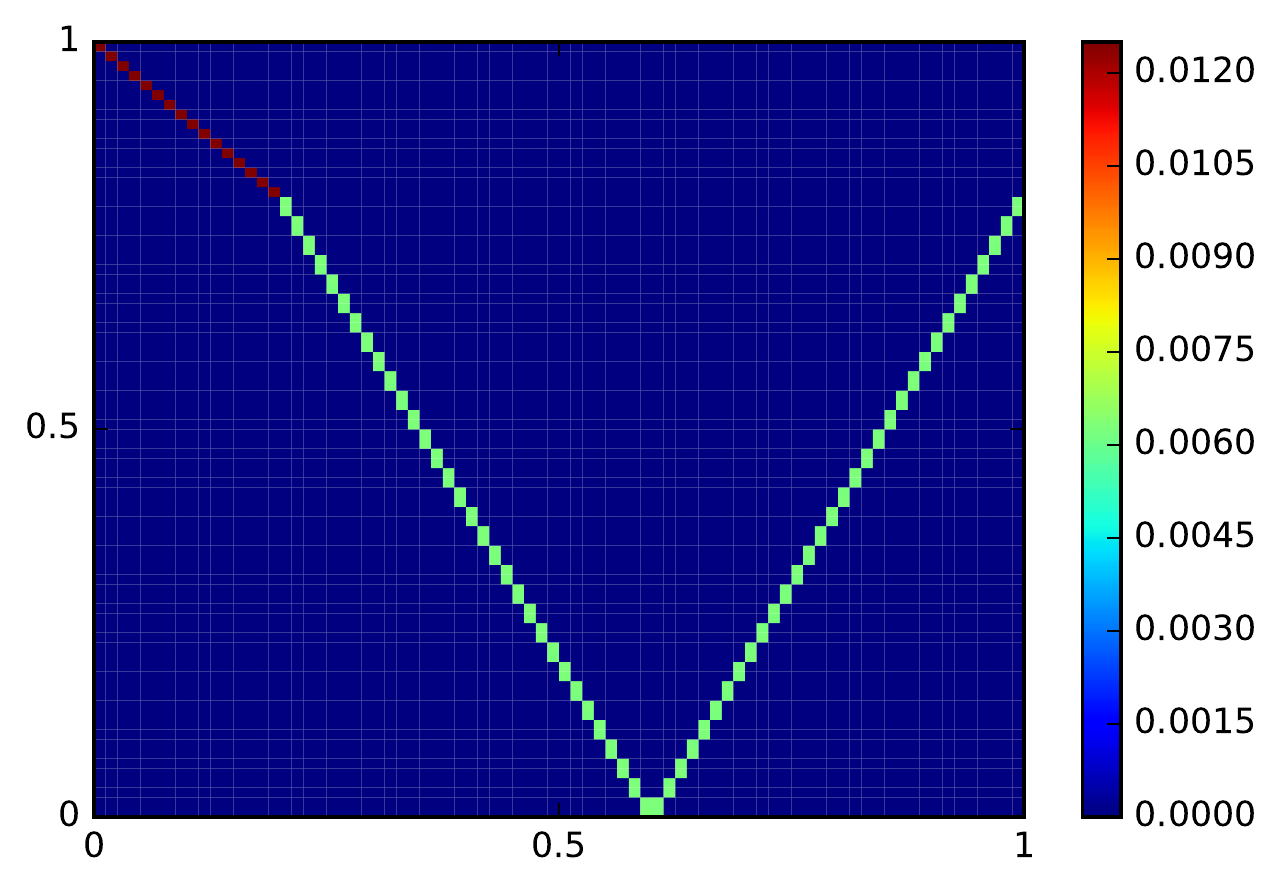}
\end{center}
\caption{4 copulas describing the dependence between $X \sim \mathcal{U}([0,1])$ and $Y \sim (X \pm \epsilon_i)^2$, where $\epsilon_i$ is a constant noise specific for each distribution. $X$ and $Y$ are counter-monotonic (more or less) half of the time, and co-monotonic (more or less) half of the time}
\label{fig:noise_patterns}
\end{figure}

\begin{figure}
\begin{center}
\includegraphics[width=0.48\linewidth]{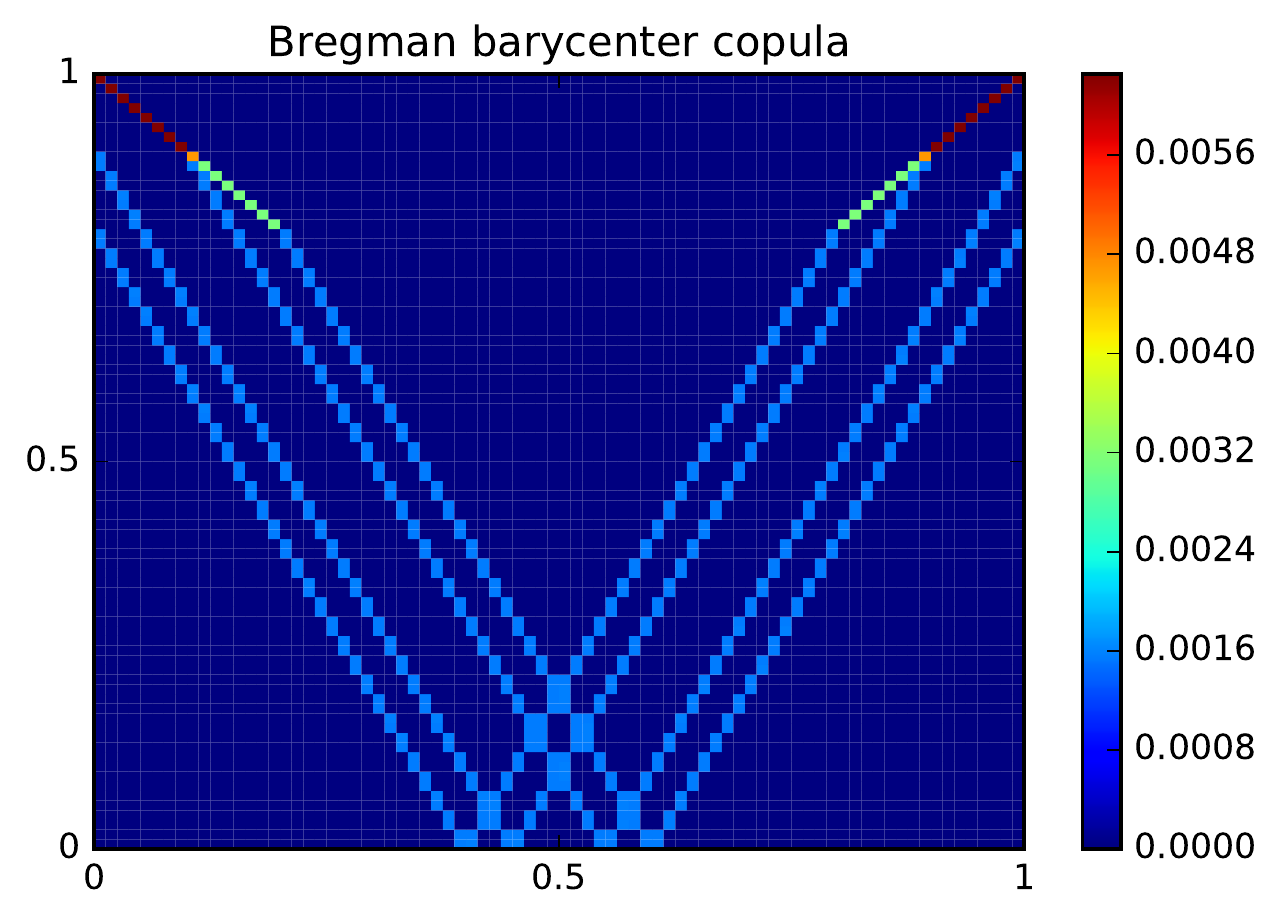}
\includegraphics[width=0.48\linewidth]{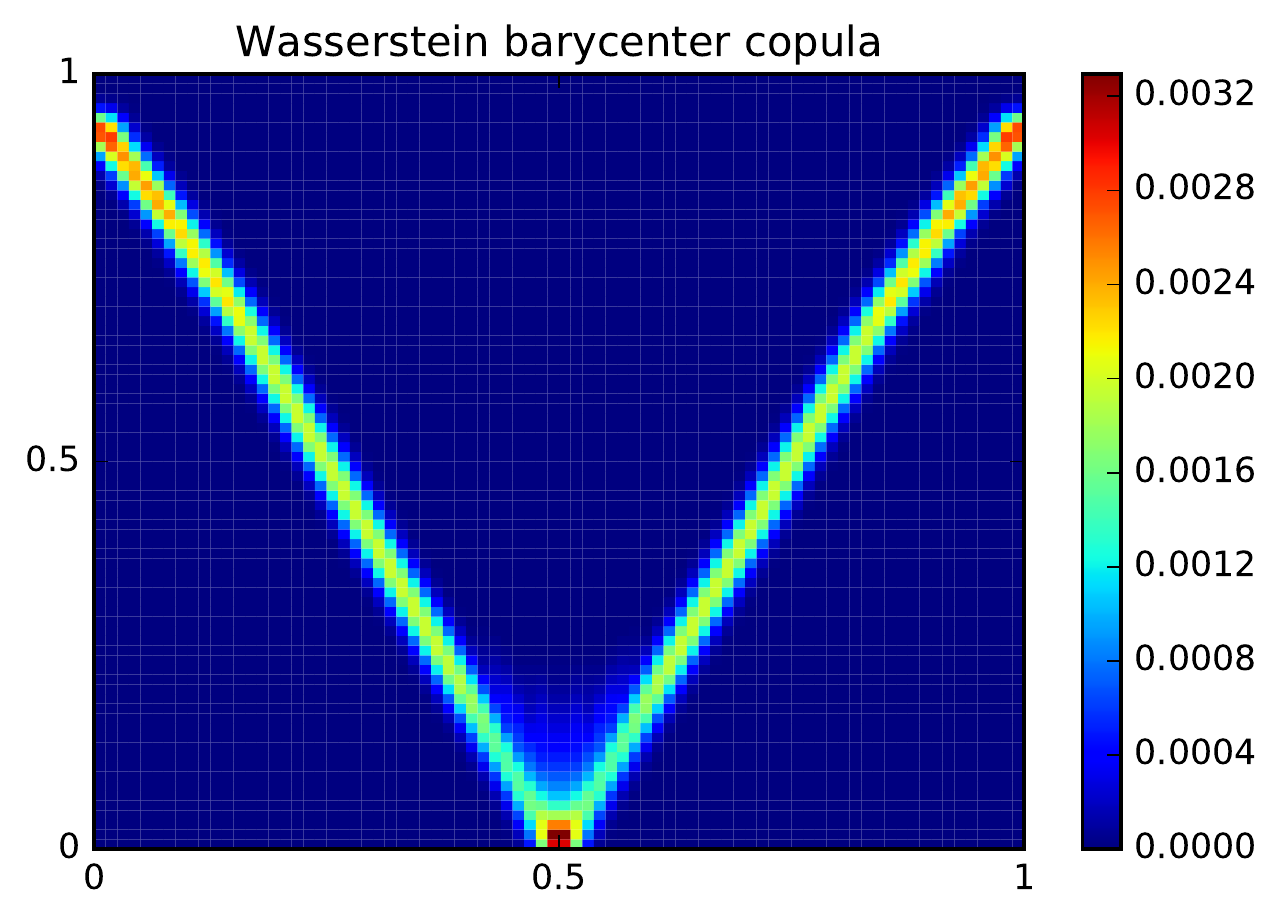}
\end{center}
\caption{Barycenter of the 4 copulas from Figure~\ref{fig:noise_patterns} for: (left) Bregman geometry \cite{banerjee2005clustering} (which includes, for example, squared Euclidean and Kullback-Leibler distances); (right) Wasserstein geometry. 
Notice that the Wasserstein barycenter better describes the underlying dependence between $X$ and $Y$: the copula encodes a functional association.
This is not the case for the Bregman barycenter}
\label{fig:barycenters}
\end{figure}

Assume we have $N$ variables $X_1,\ldots,X_N$, and $T$ observations for each of them.
First, we compute $\binom{N}{2}=O(N^2)$ empirical copulas which represent 
the dependence structure between all the couples ($X_i,X_j)$.
Then, we summarize all these distributions using a center-based clustering algorithm, 
and extract the clusters centers
using a fast computation of Wasserstein barycenters \cite{DBLP:conf/icml/CuturiD14}.
A given center represents the mean dependence between the couples $(X_i,X_j)$ inside the corresponding cluster.
Figure~\ref{fig:noise_patterns} and~\ref{fig:barycenters} illustrate
why a Wasserstein $W_2$ barycenter, i.e. the minimizer $\mu^\star$ of $\frac{1}{N} \sum_{i=1}^N W_2^{2}(\mu,\nu_i)$ \cite{agueh2011barycenters} where $\{\nu_1,\ldots,\nu_N\}$ is a
set of $N$ measures (here, bivariate empirical copulas), is more relevant to our needs:
we benefit from robustness against small deformations of the dependence patterns.

\textbf{Example.} In Table~\ref{tab:gt}, we display some interesting dependence patterns 
which can be found in UCI datasets \url{http://archive.ics.uci.edu/ml/}. In this case, variables $X_1,\ldots,X_N$ are the $N$ features.
Some associations are easy to explain (e.g. top left copula representing the relation 
between radius and area of roughly round cells in the \texttt{Breast Cancer Wisconsin (Diagnostic) Data Set})
whereas some others less (e.g. top row third copula from the left which represents the relation between the perimeter and the fractal dimension
of the cells).

An equitable copula-based dependence measure such as the one described in \cite{ghahramani2012copula}
may detect them well, but will also detect the spurious ones which are due to artifacts in the data (or pure chance).
With this approach, one can spot them and add them to the set of forget-dependence copulas.
For these reasons, we think that this approach could improve the feature selection correlation-based 
approaches \cite{hall2000correlation,yu2003feature} which rely on the hypothesis that
\textit{good feature subsets contain features highly correlated with the class, yet uncorrelated with each other \cite{hall2000correlation}}.

\begin{table*}[ht]
\centering
\caption{Dependence patterns (= clustering centroids) found between variables in UCI datasets}
\centering
\begin{tabular}{lccccc}
\hline
Breast Cancer (wdbc) & 
\includegraphics[width=0.1\textwidth]{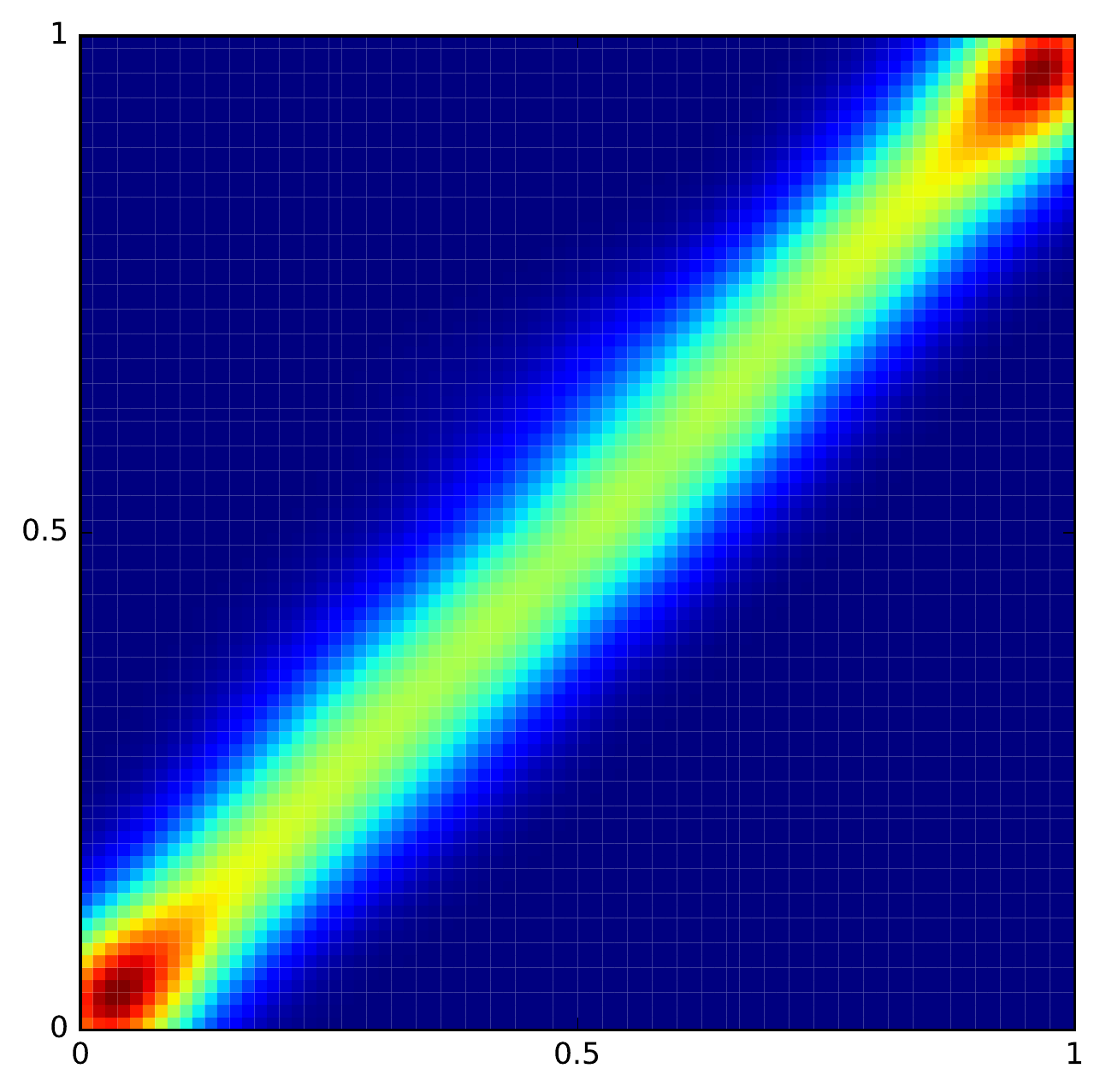} & 
\includegraphics[width=0.1\textwidth]{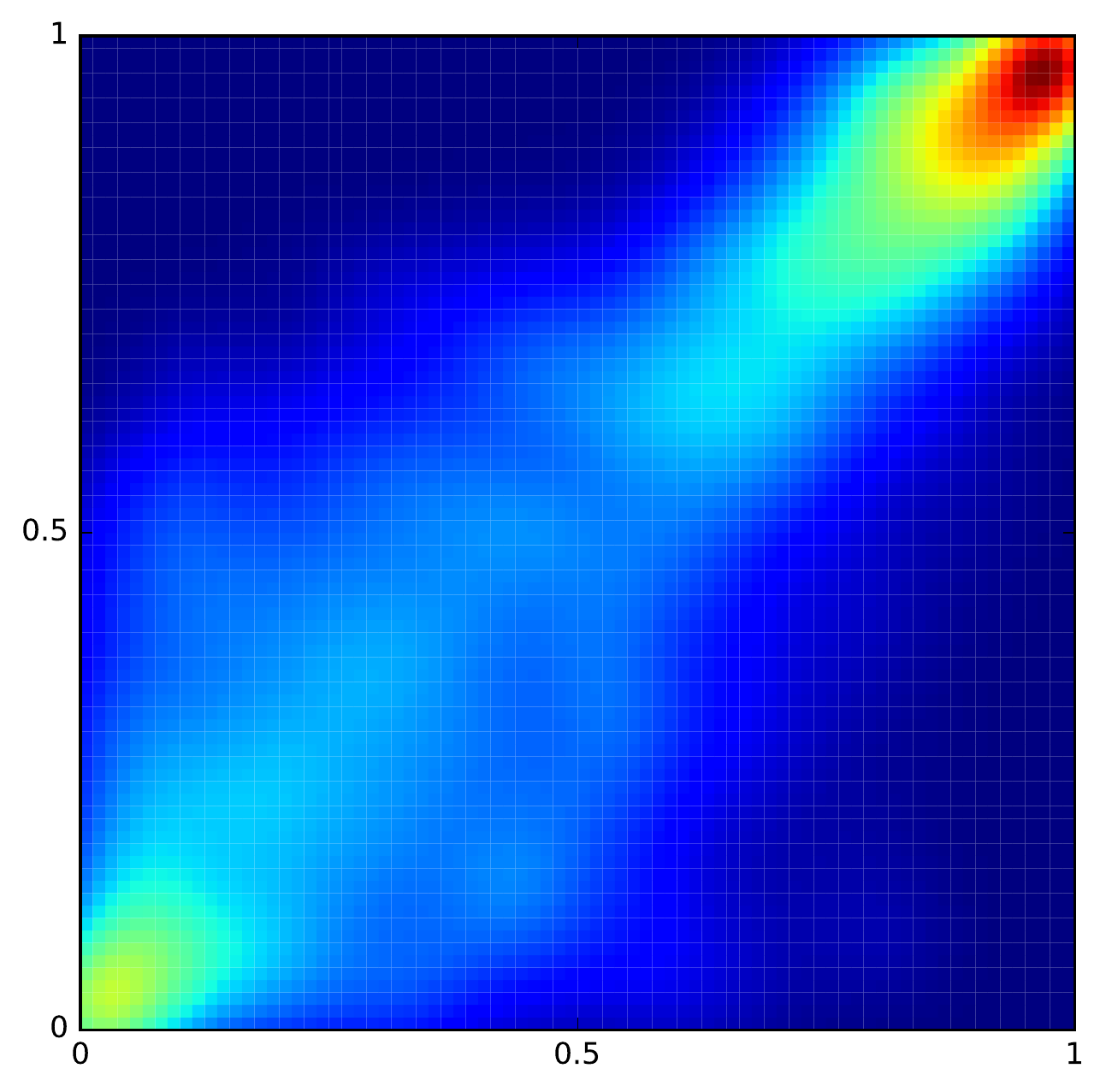} &
\includegraphics[width=0.1\textwidth]{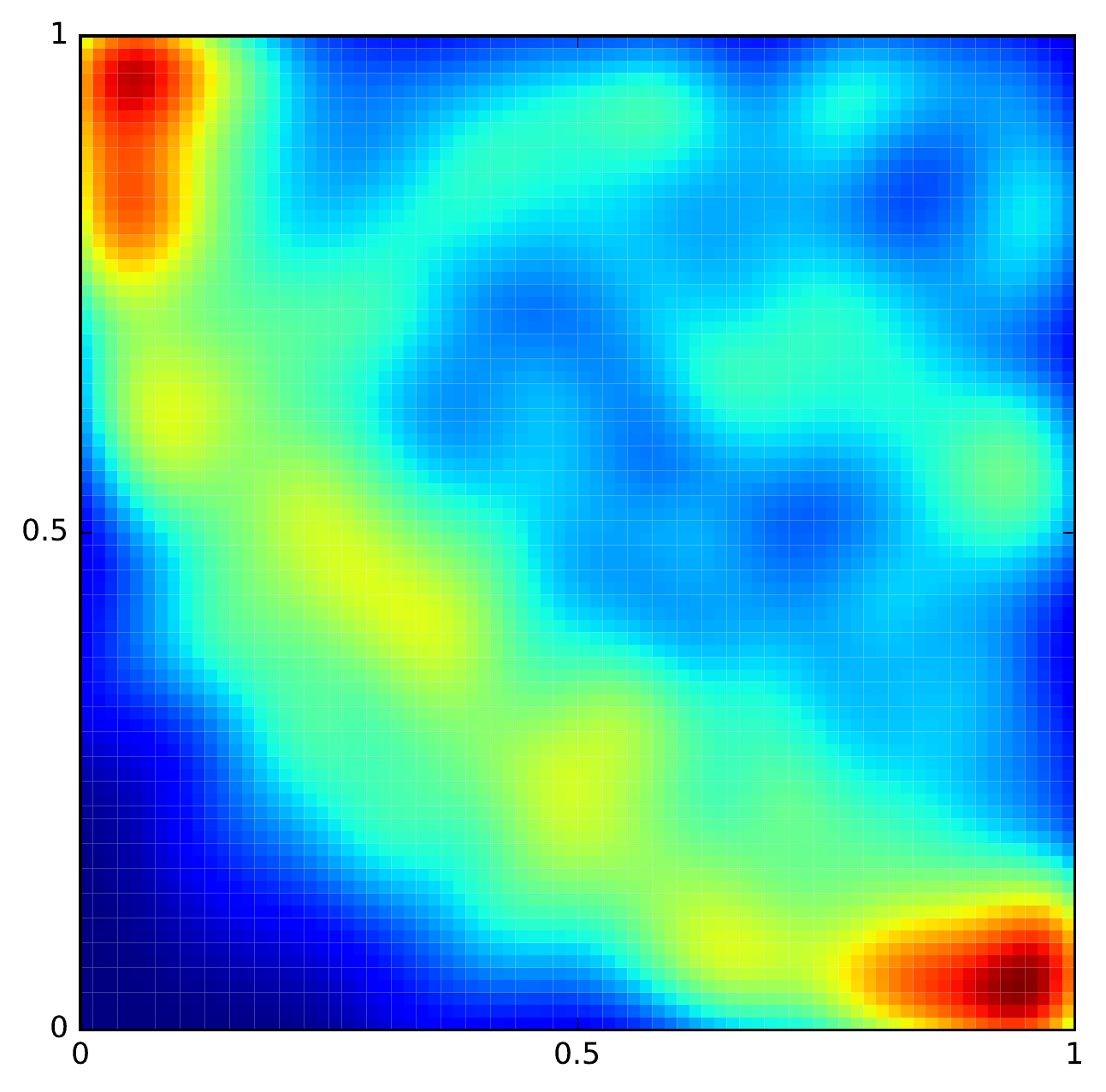} & 
\includegraphics[width=0.1\textwidth]{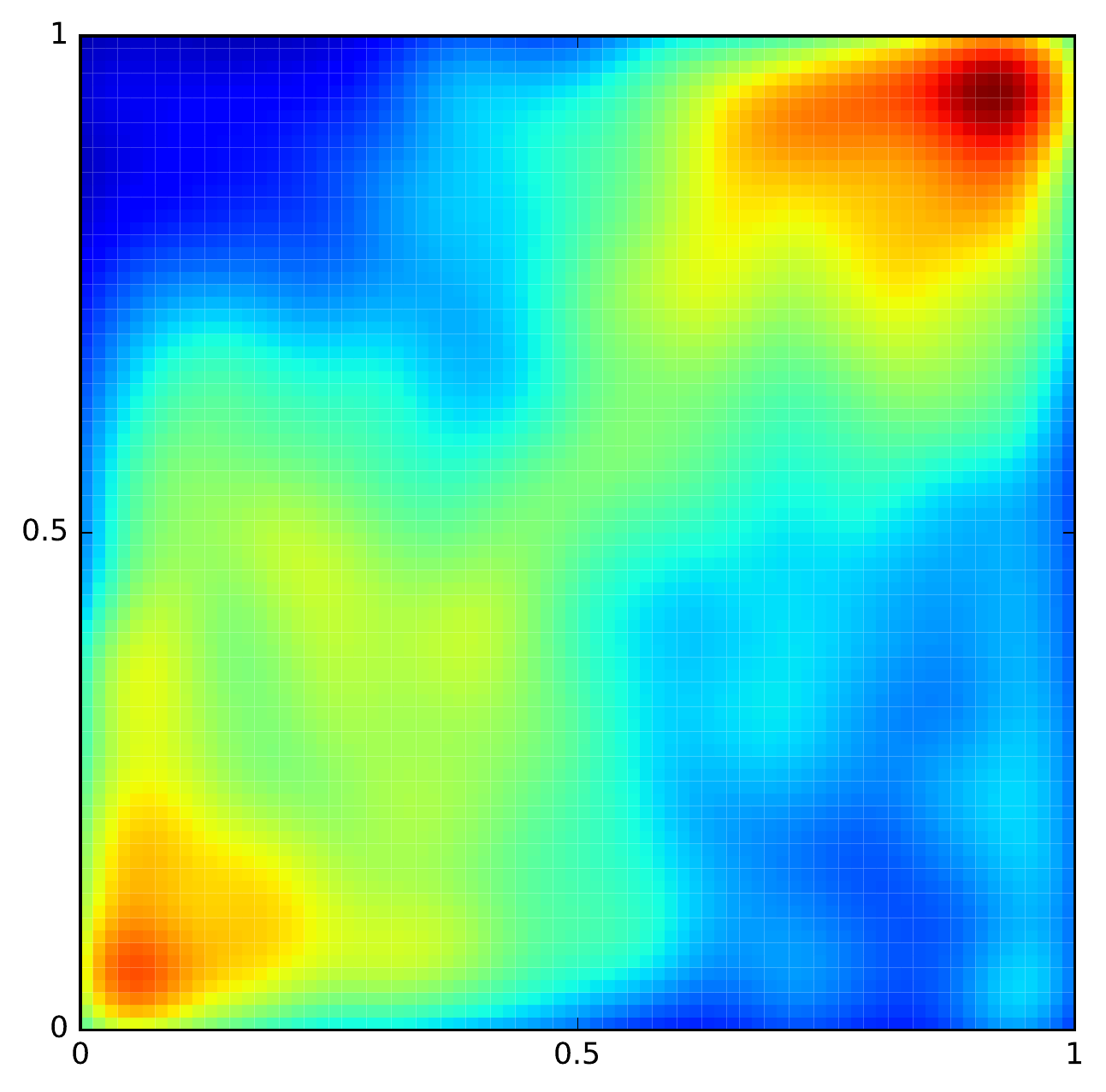} &
\includegraphics[width=0.1\textwidth]{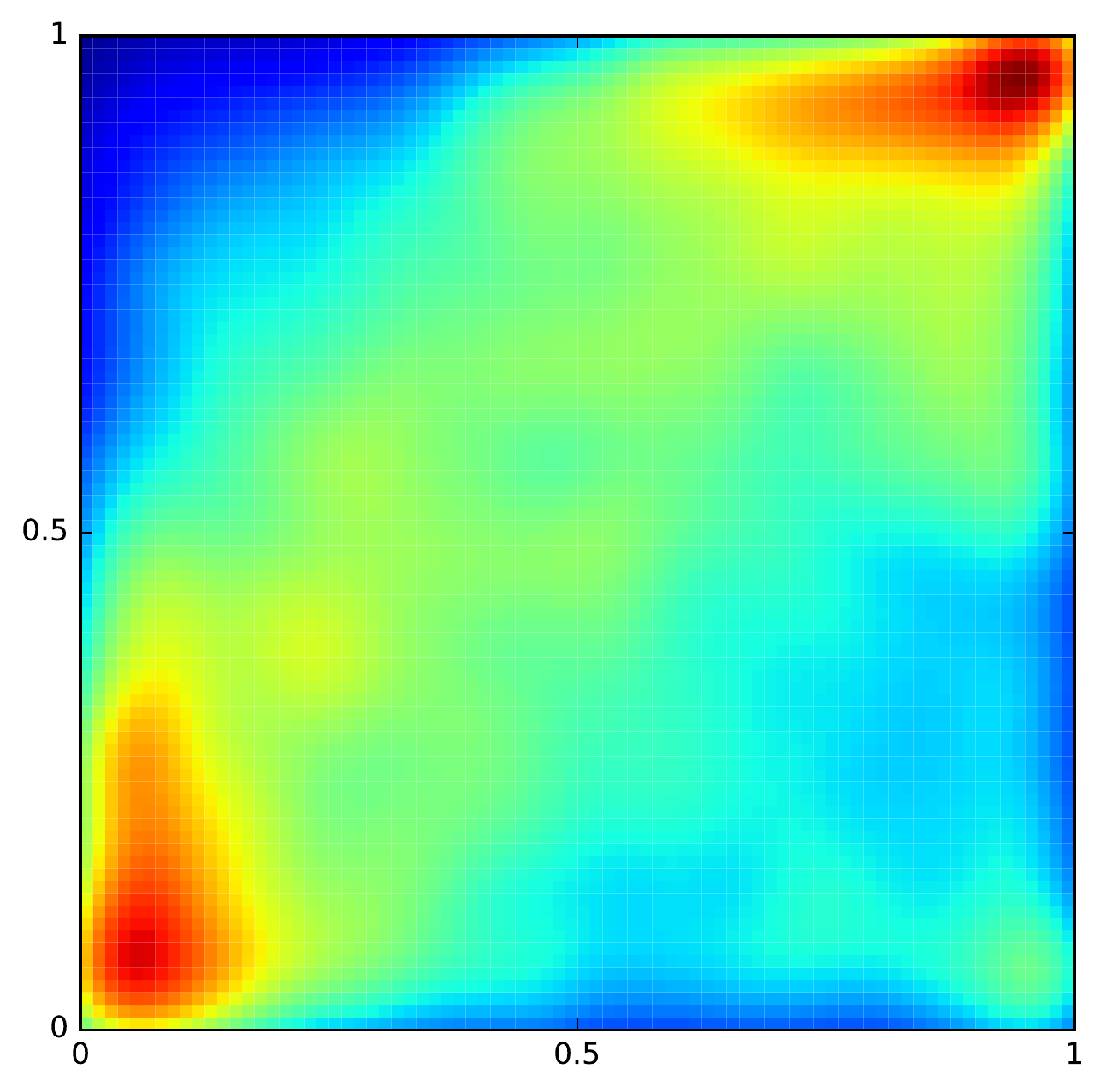}\\
\hline
Libras Movement & \includegraphics[width=0.1\textwidth]{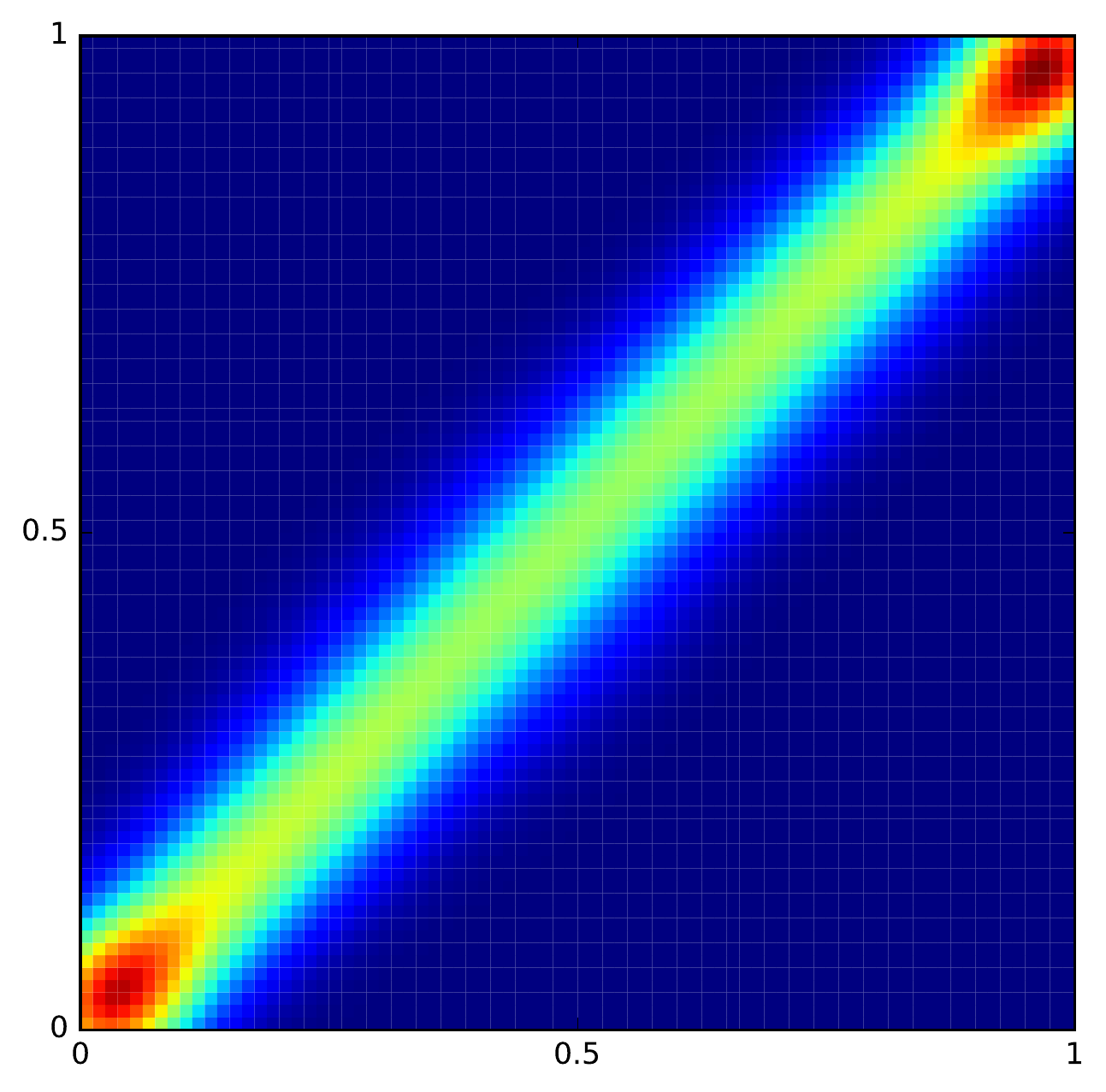} & 
\includegraphics[width=0.1\textwidth]{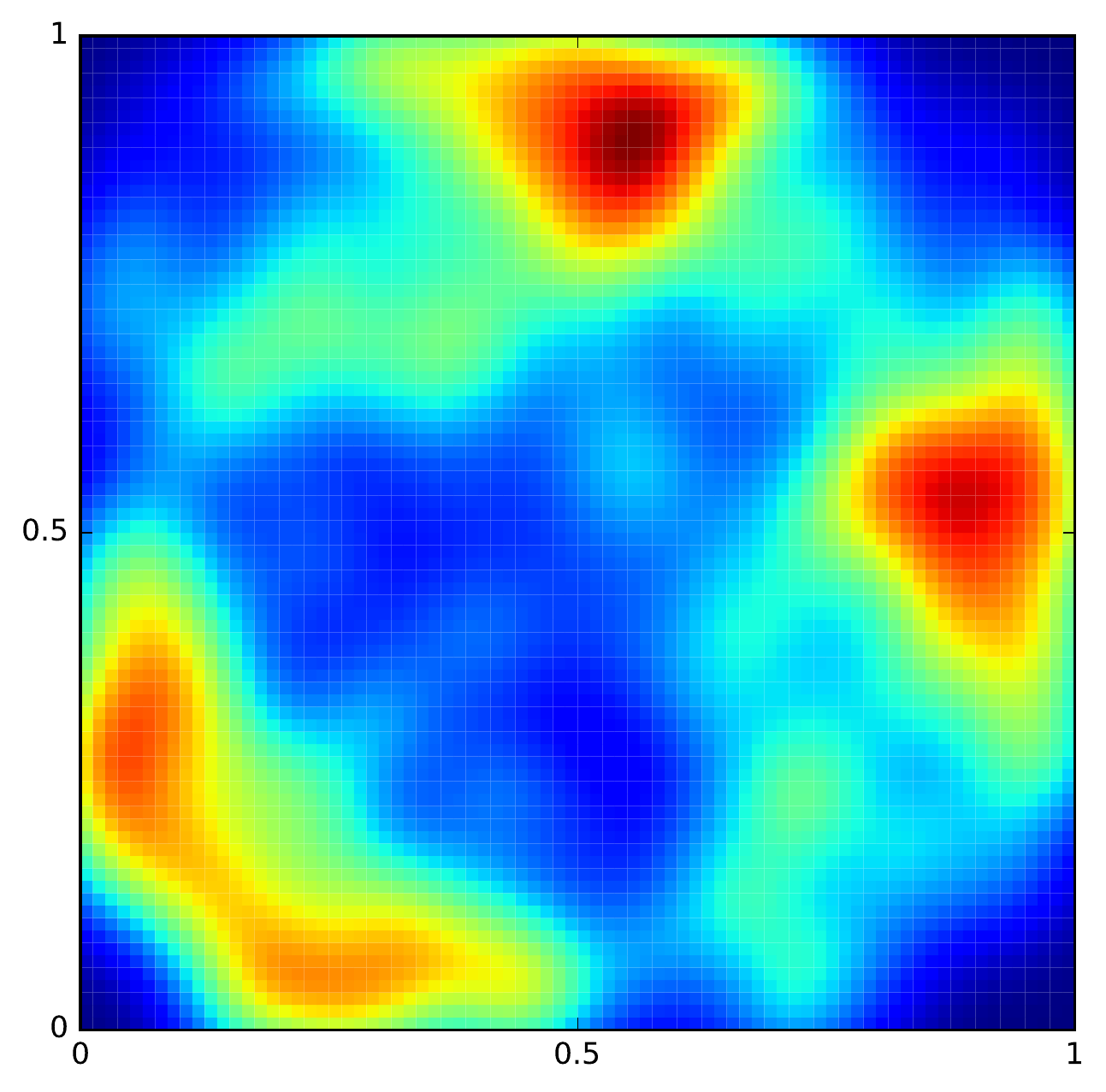} &
\includegraphics[width=0.1\textwidth]{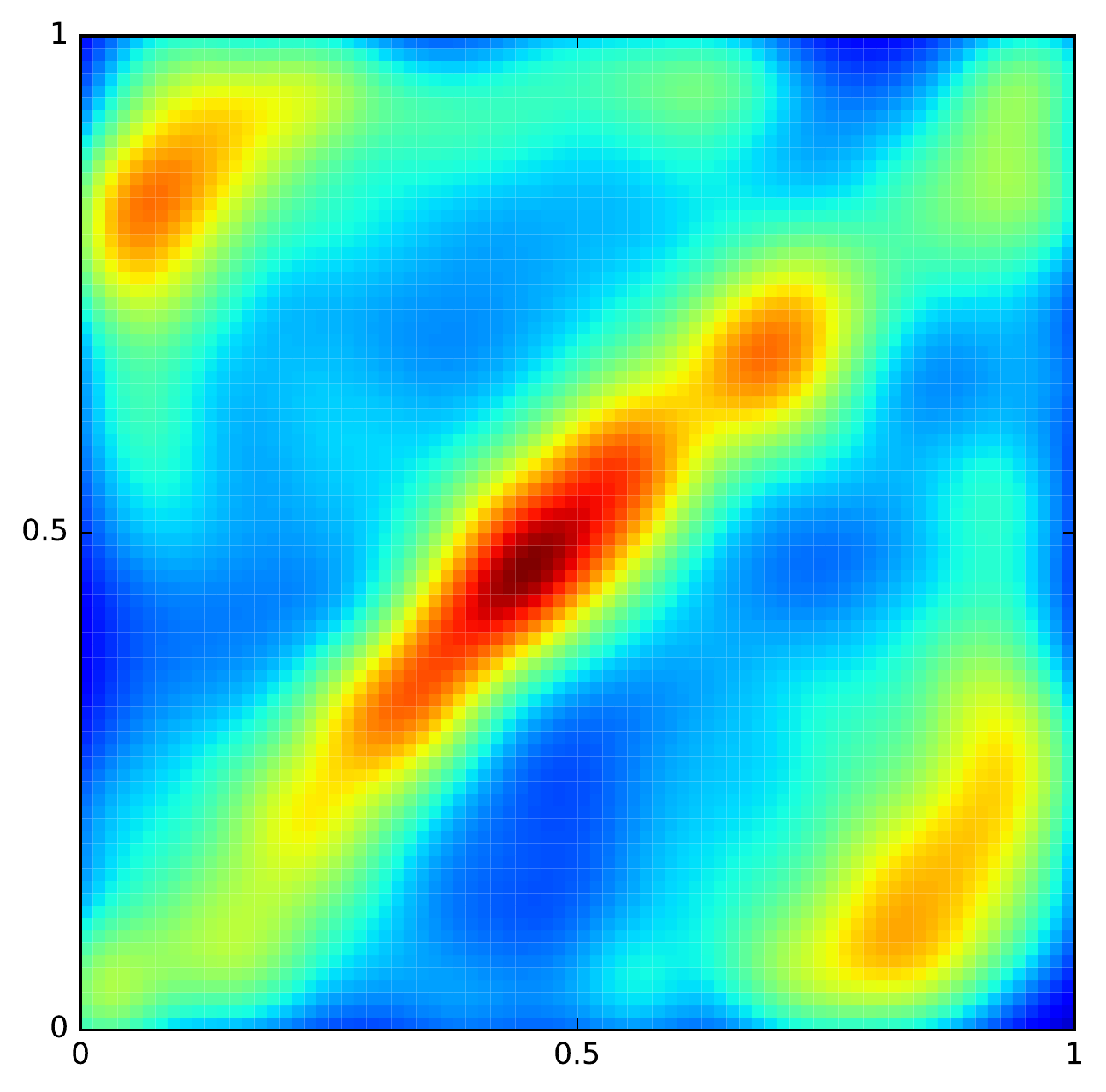} & 
\includegraphics[width=0.1\textwidth]{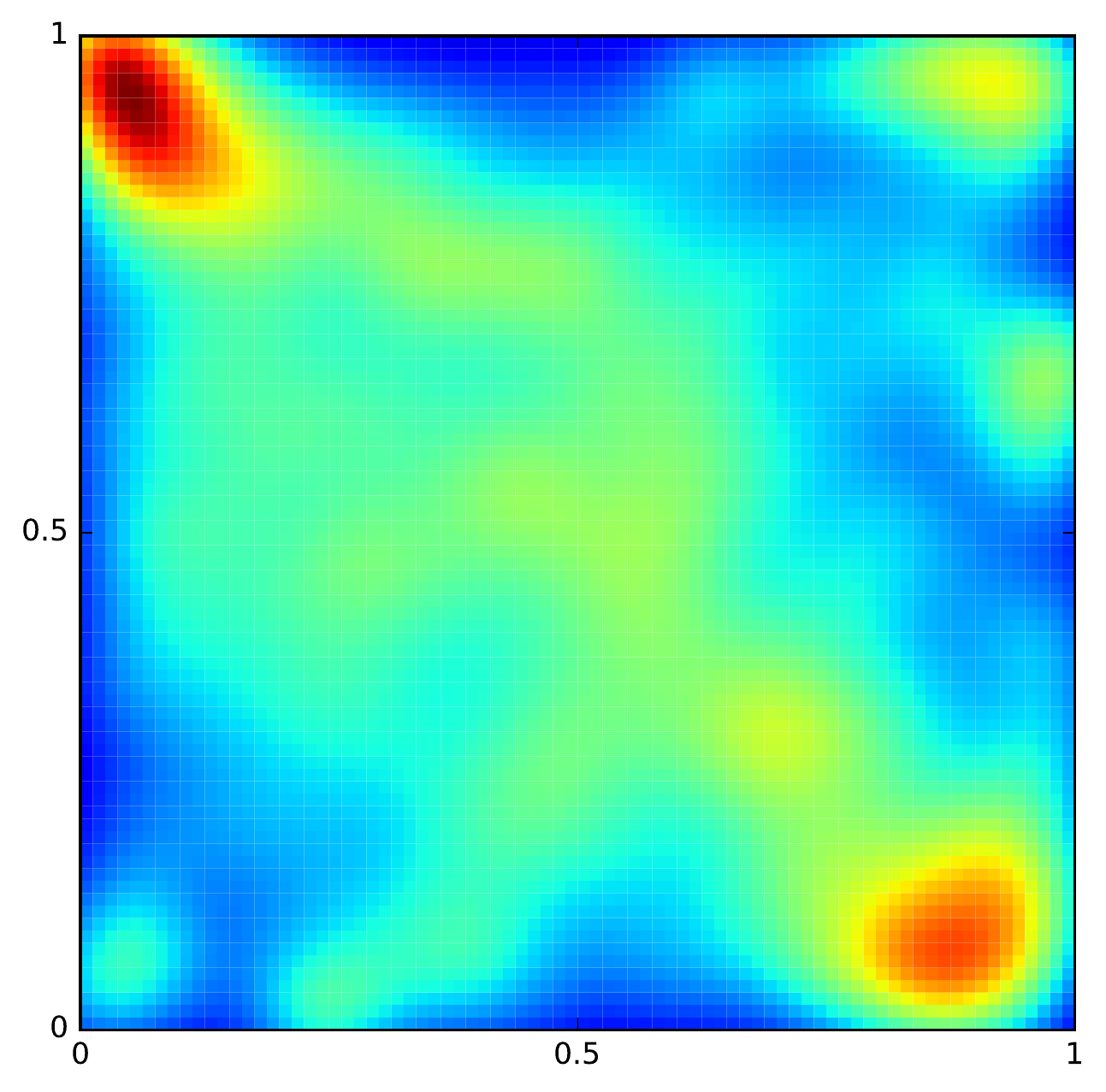} & 
\includegraphics[width=0.1\textwidth]{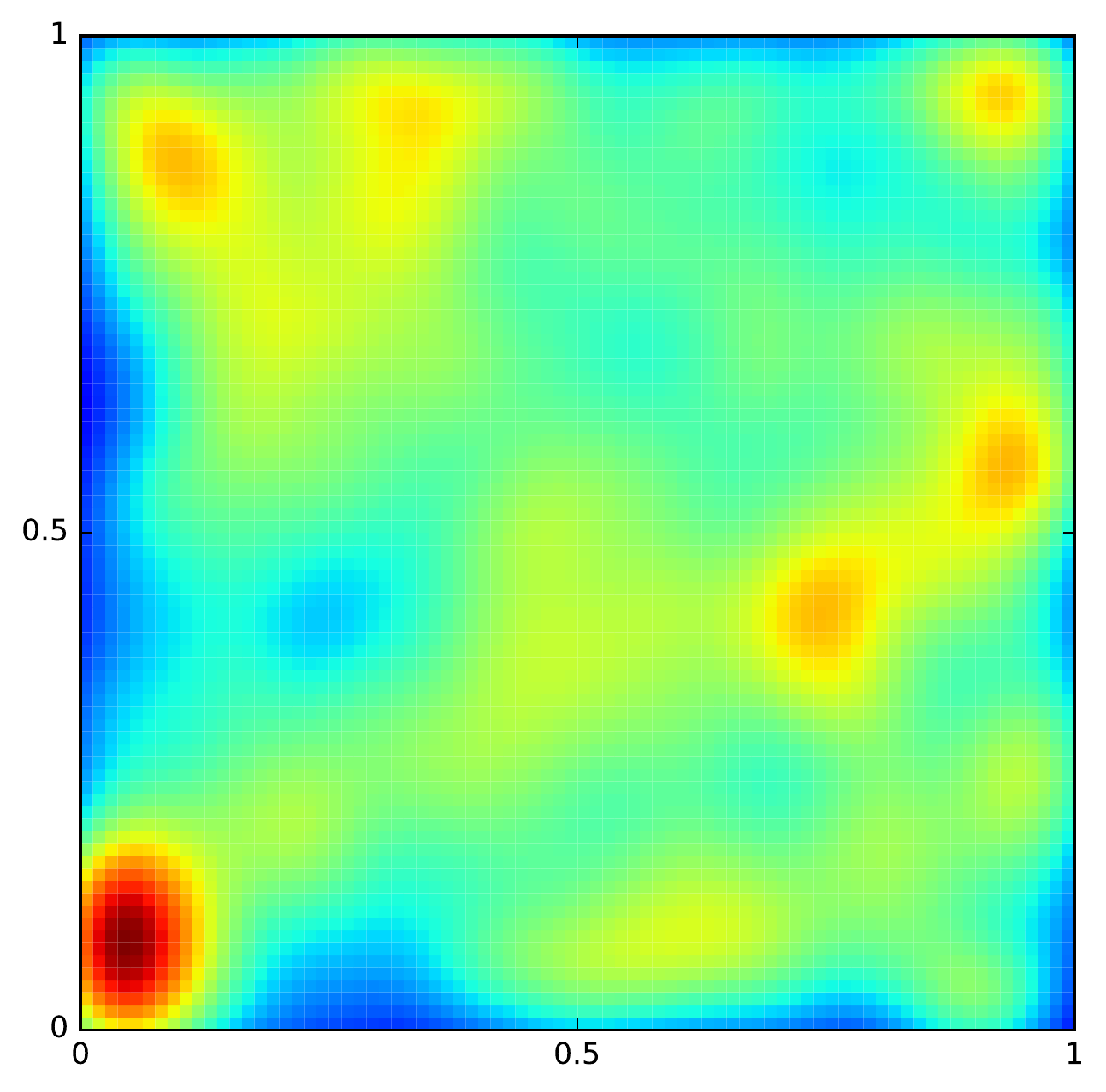}\\
\hline
Parkinsons & 
\includegraphics[width=0.1\textwidth]{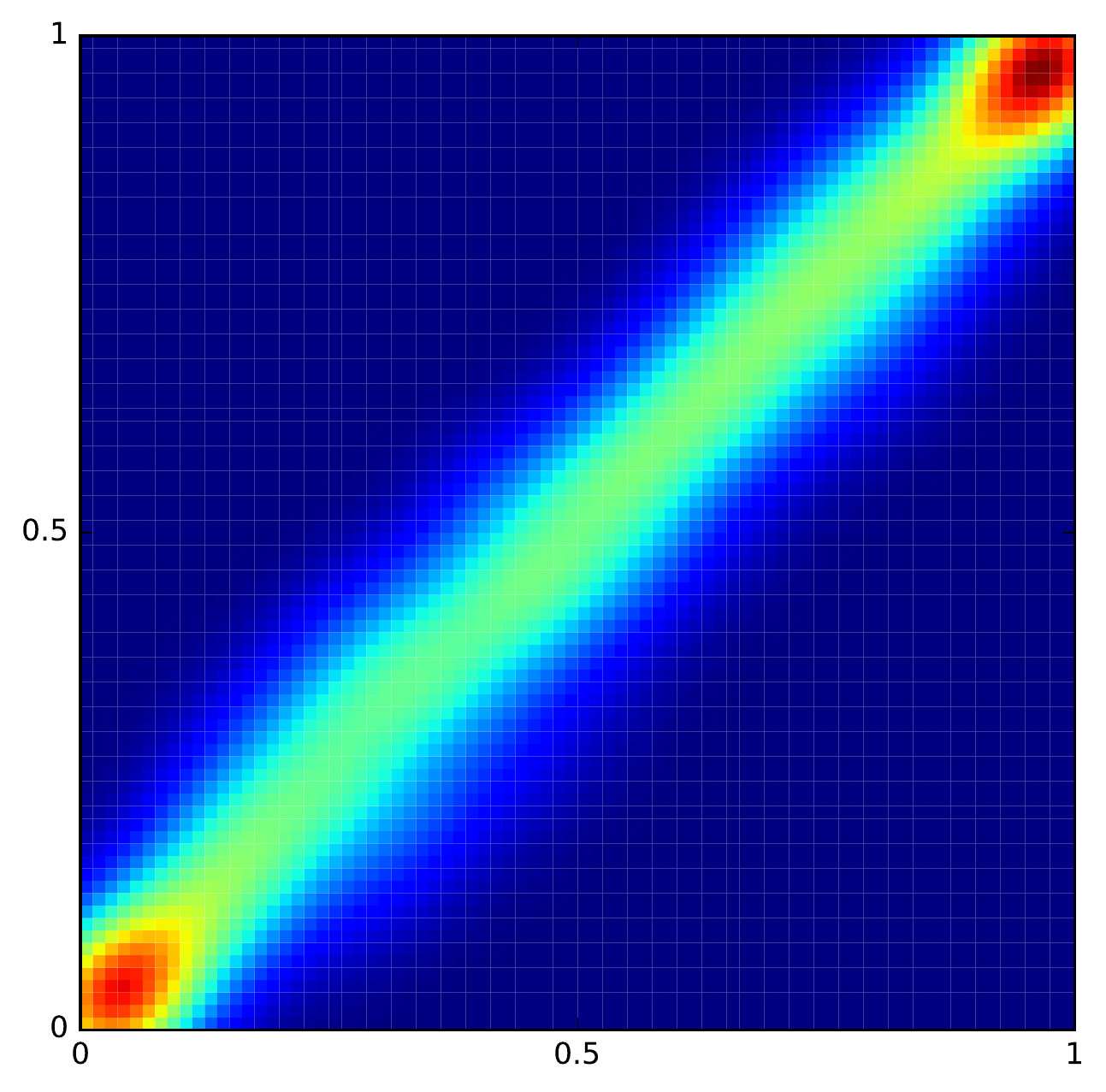} & 
\includegraphics[width=0.1\textwidth]{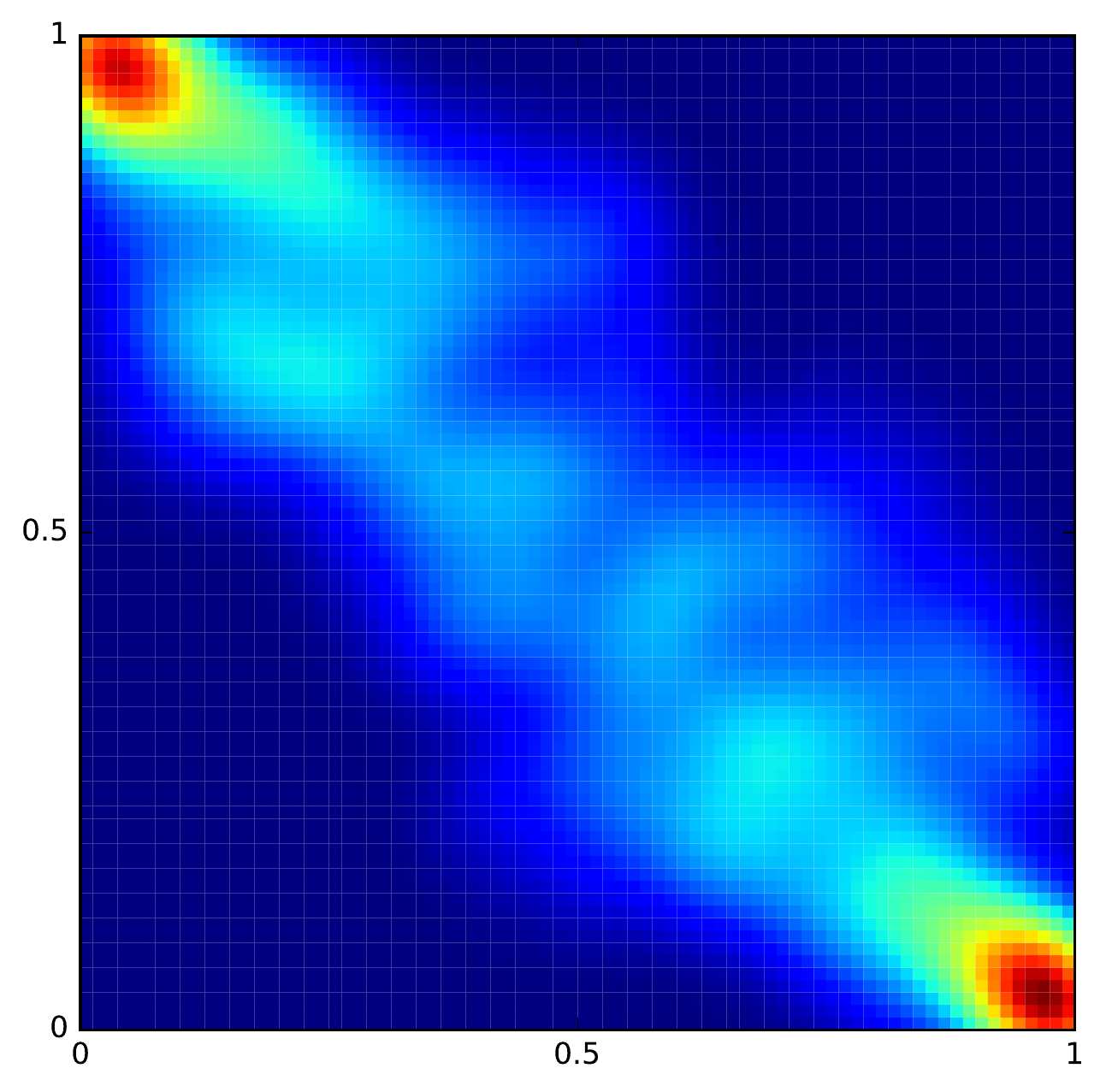} & 
\includegraphics[width=0.1\textwidth]{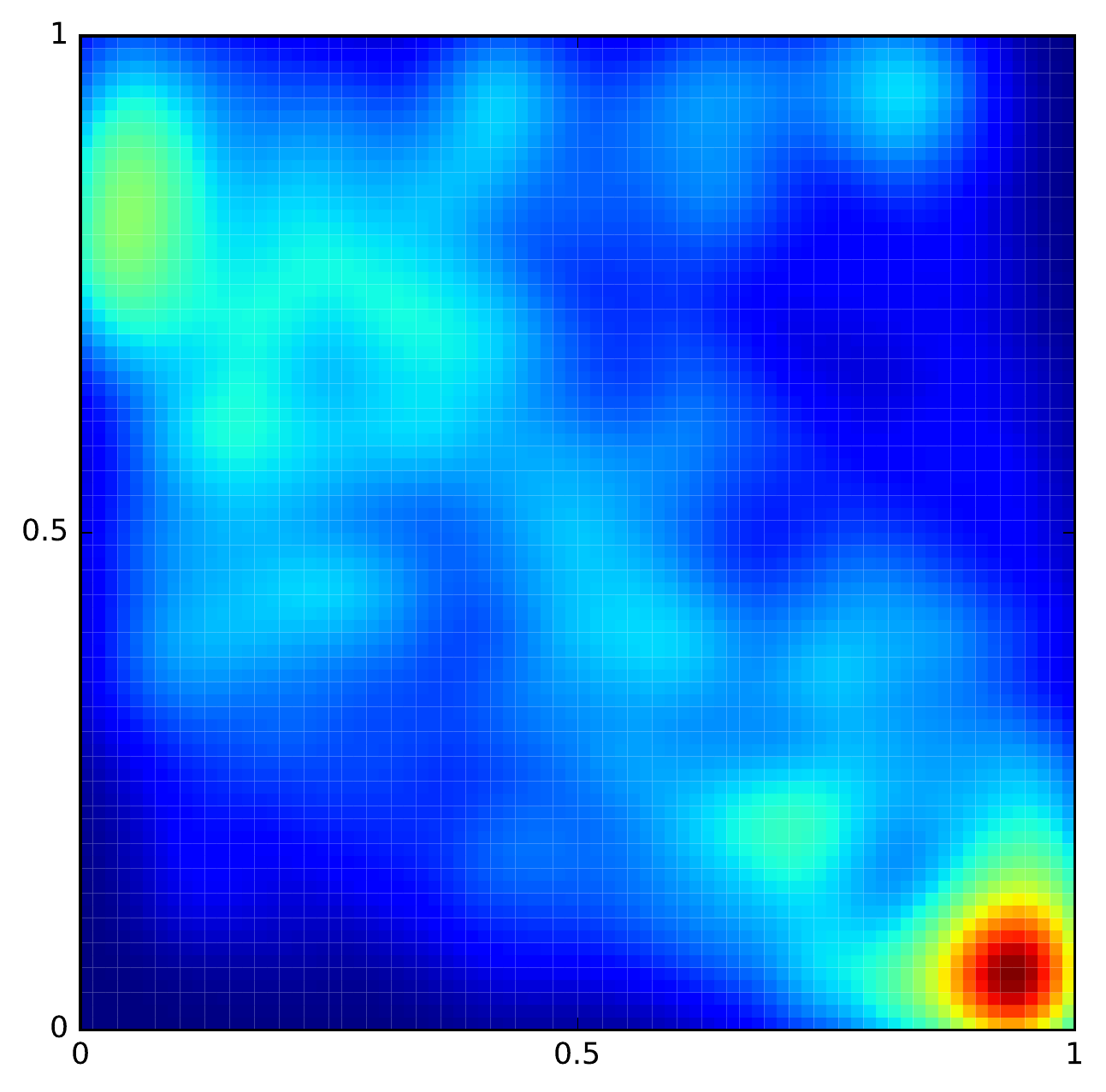} & 
\includegraphics[width=0.1\textwidth]{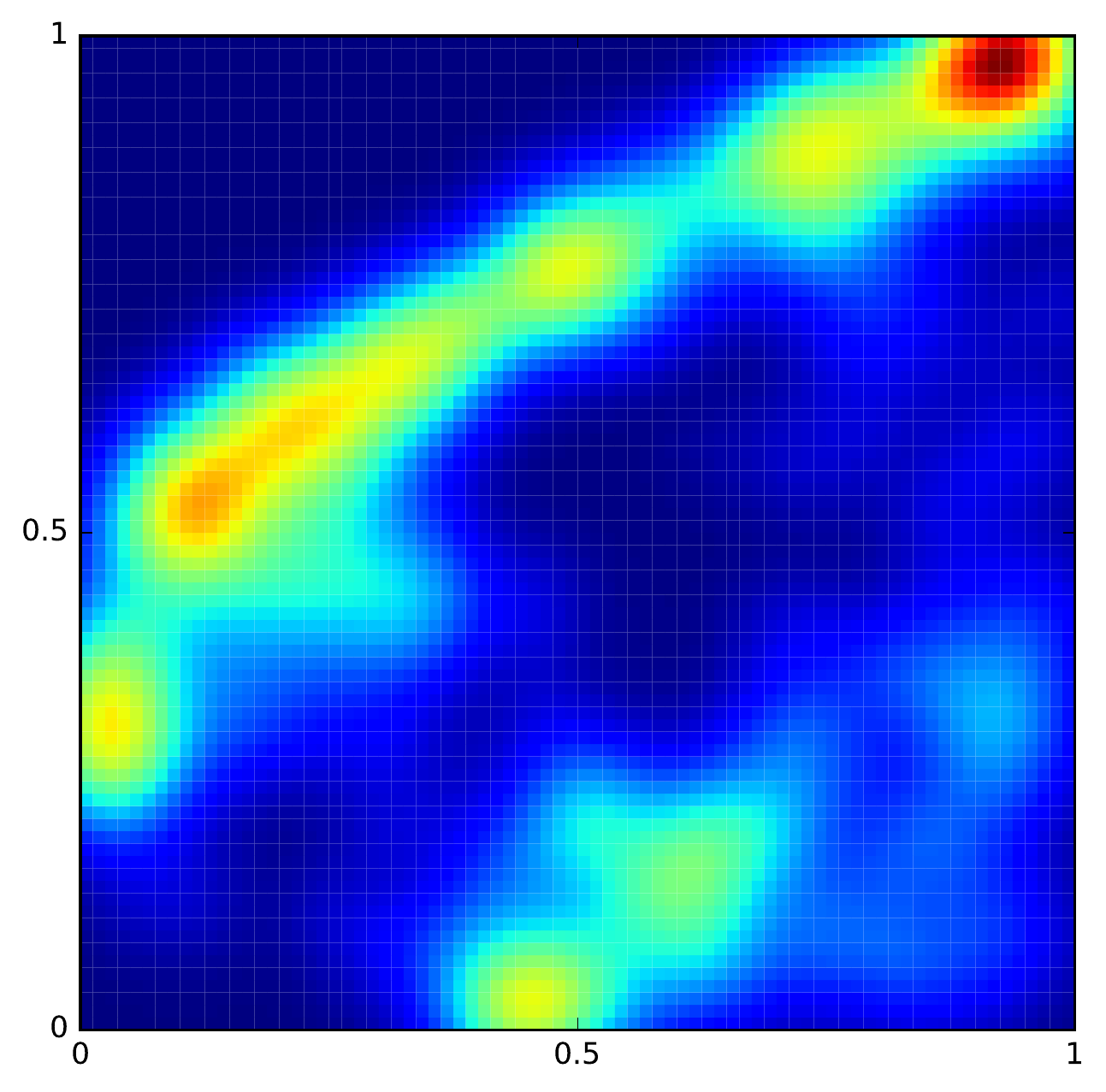} &
\includegraphics[width=0.1\textwidth]{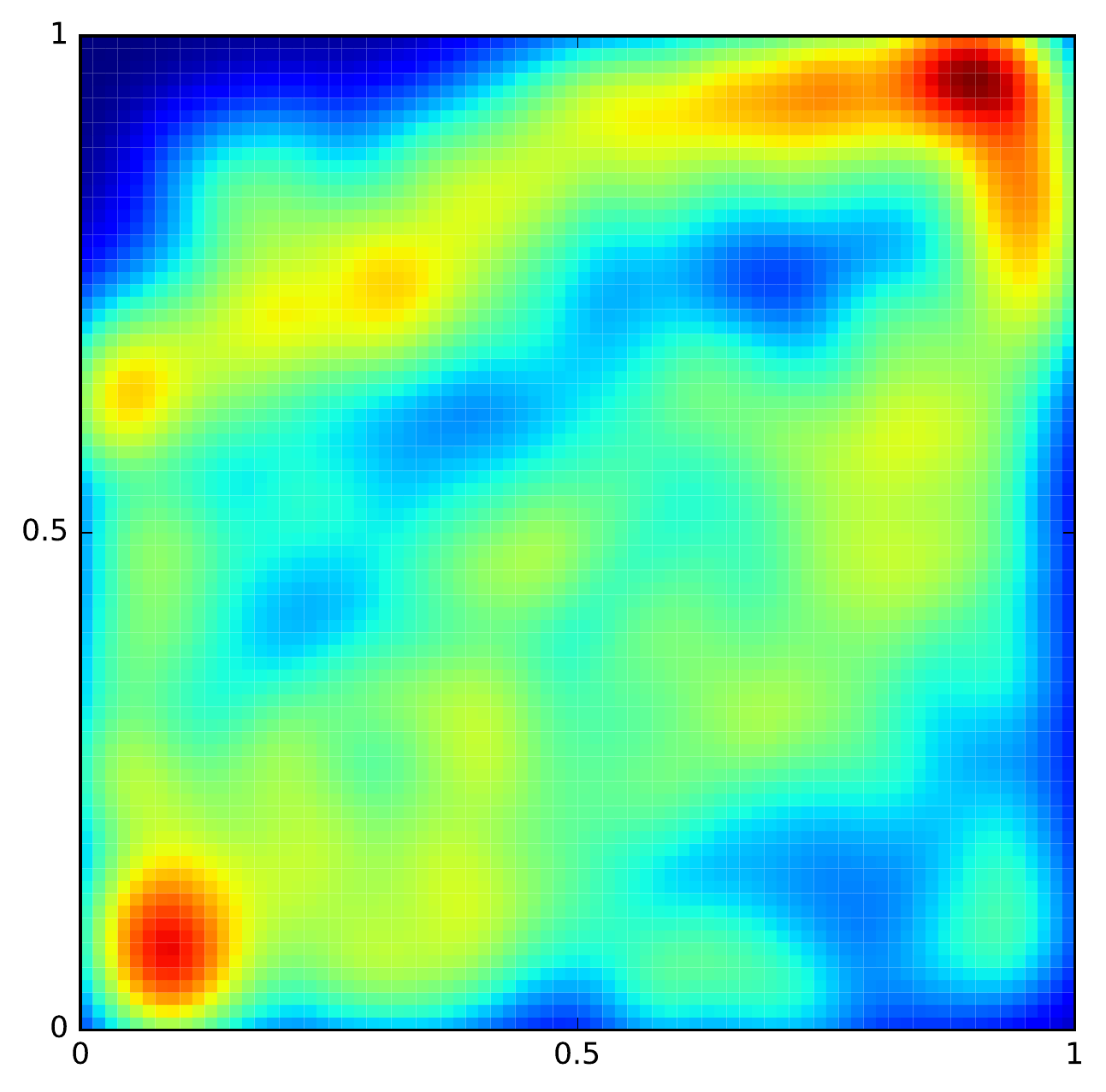}\\
\hline
Gamma Telescope & 
\includegraphics[width=0.1\textwidth]{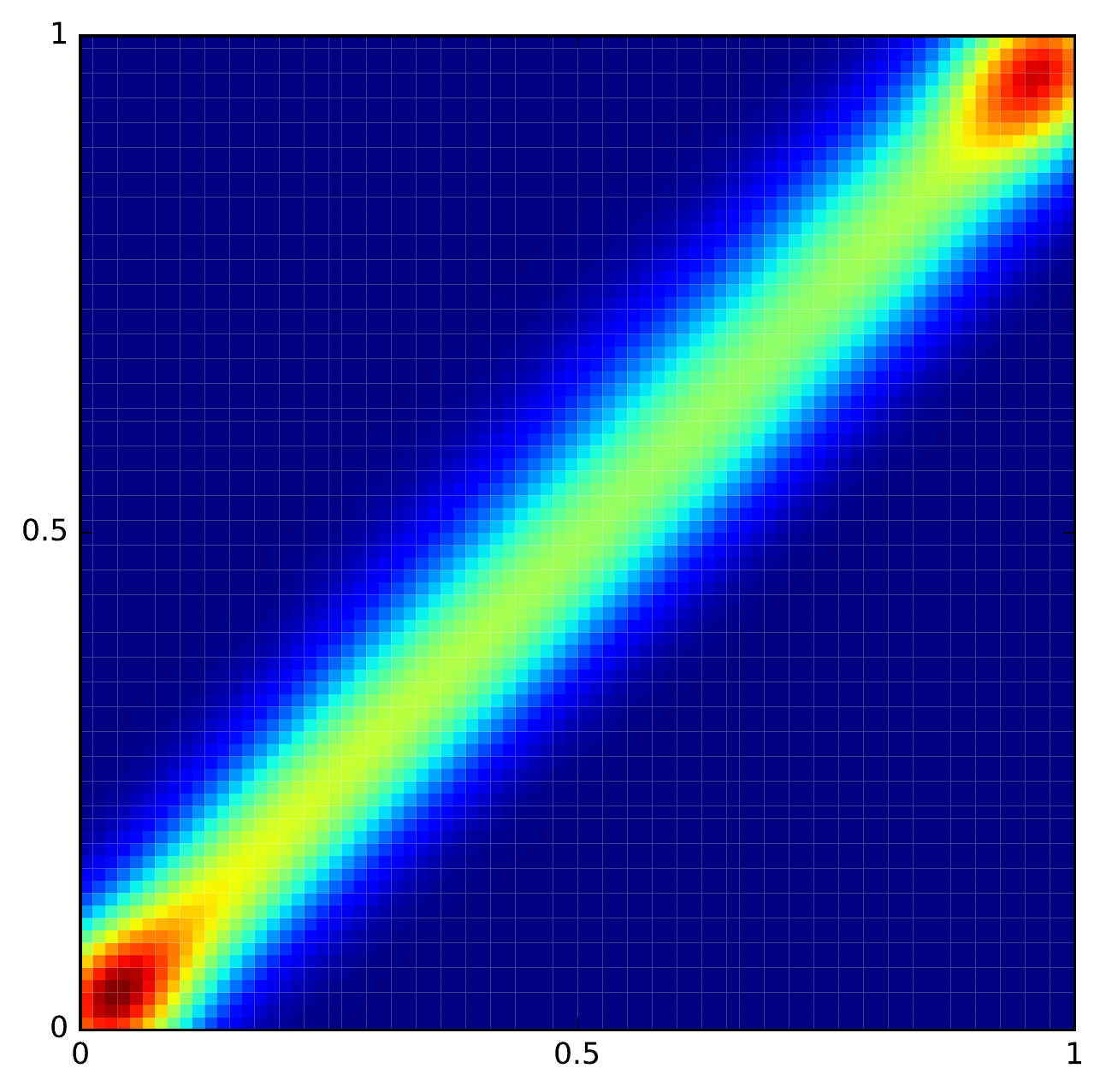} & 
\includegraphics[width=0.1\textwidth]{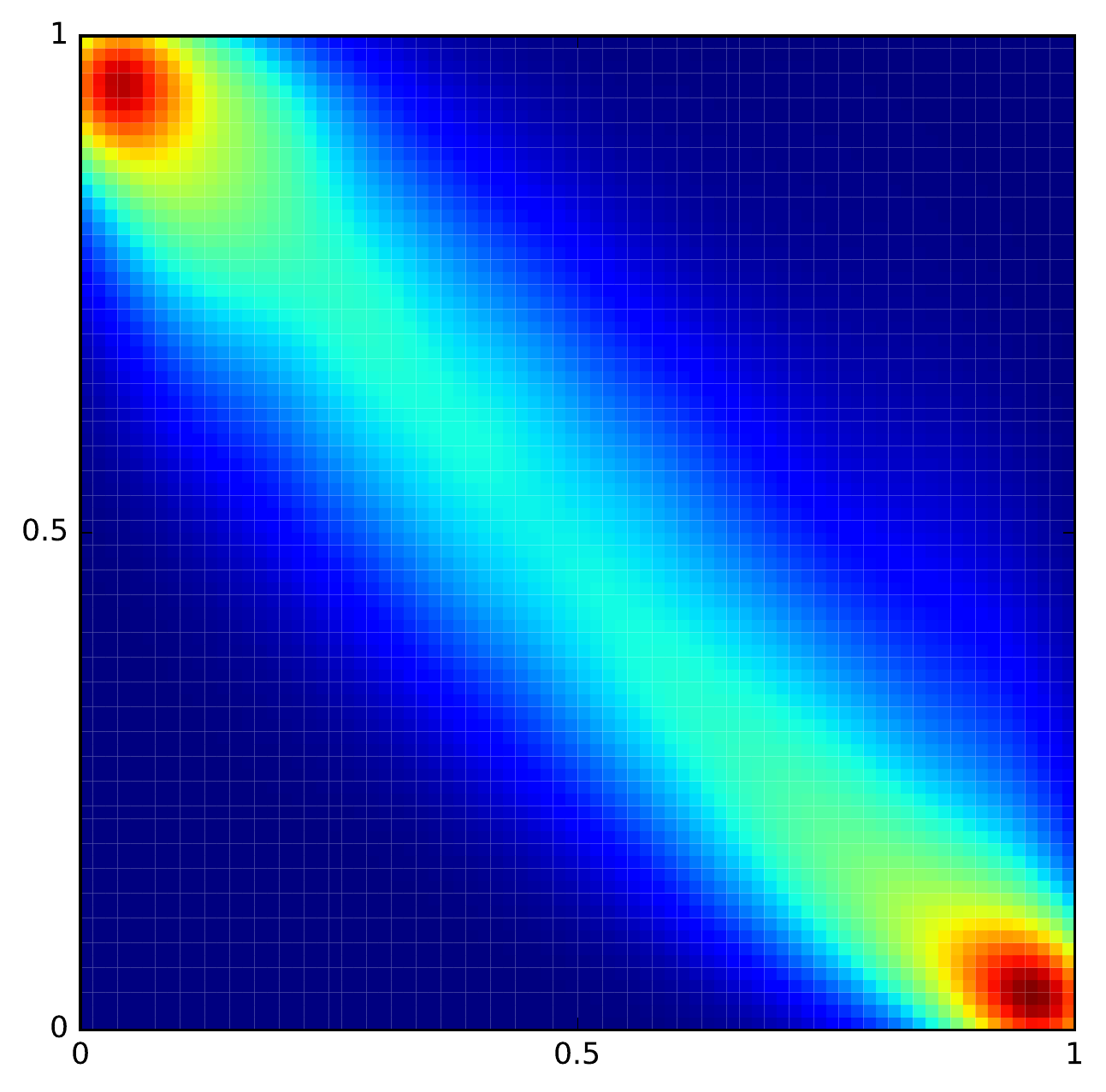} &
\includegraphics[width=0.1\textwidth]{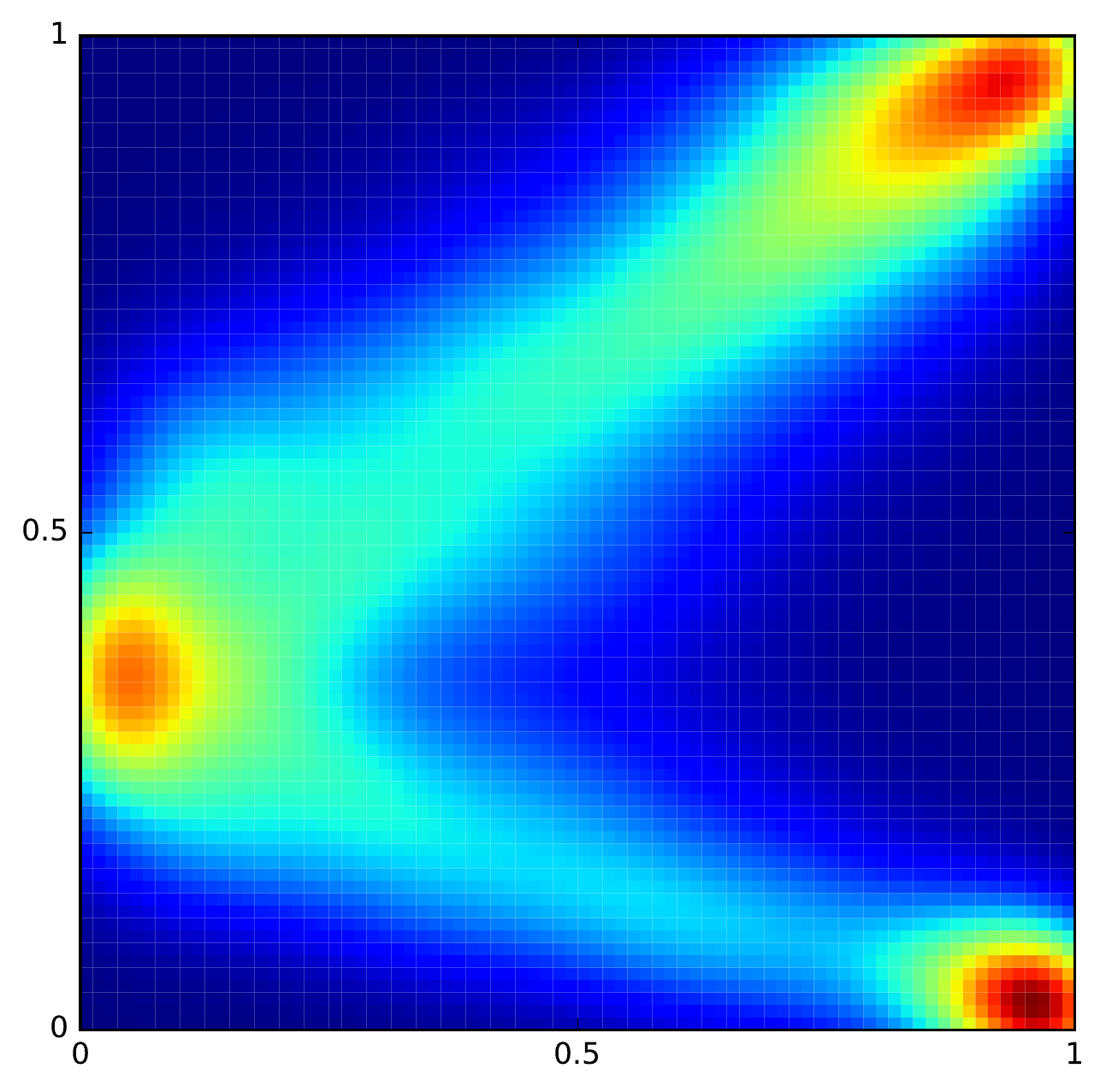} & 
\includegraphics[width=0.1\textwidth]{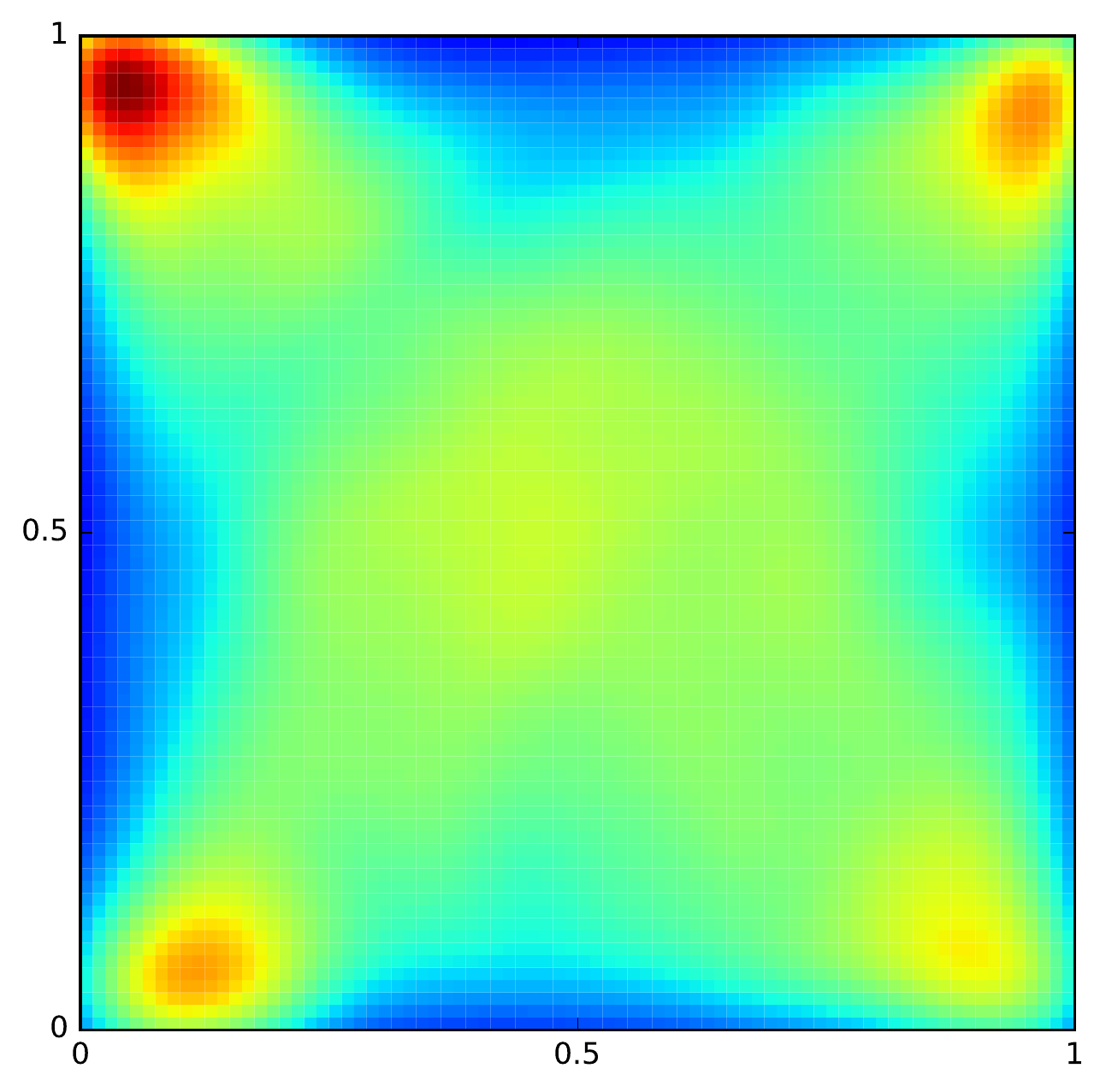} & 
\includegraphics[width=0.1\textwidth]{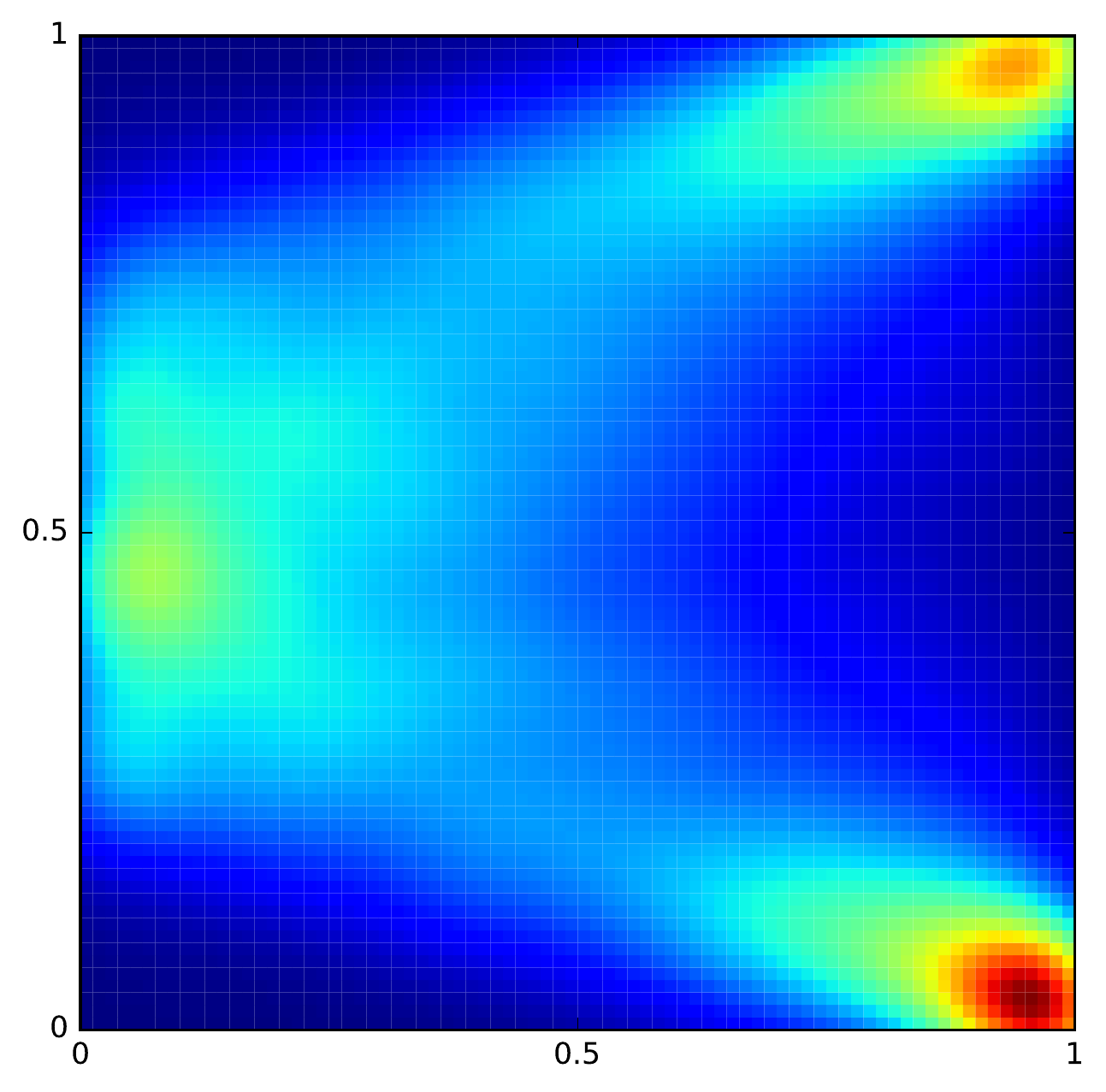}\\
\hline
\end{tabular}
\label{tab:gt}
\end{table*}



\subsubsection{Targets as hypotheses from an expert}

One can specify dependence hypotheses, generate the corresponding copulas,
then measure and rank correlations with respect to them.
For example, one can answer to questions such as:
Which are the pairs of assets that are usually positively correlated 
for small variations but uncorrelated otherwise?
In \cite{durante2009rectangular}, authors present a method for constructing bivariate copulas 
by changing the values that a given copula assumes on some subrectangles of the unit square.
They discuss some applications of their methodology including the construction of copulas
with different tail dependencies.
Building \textit{target} and \textit{forget} copulas is another one.
In the Experiments section, we illustrate its use to answer the previous question and other dependence queries.

\section{Experiments}

\subsection{Exploration of financial correlations}

We illustrate the first part of the methodology with three different datasets of financial time series.
These time series consist in the daily returns of stocks (40 stocks from the CAC 40 index comprising the French highest market capitalizations), credit default swaps (75 CDS from the iTraxx Crossover index comprising the most liquid sub-investment grade European entities) and foreign exchange rates (80 FX rates of major world currencies) between January 2006 and August 2016. We display some of the clustering centroids obtained for each asset class on the top row, and below we display their corresponding Gaussian copulas parameterized by the estimated linear correlations. 
Notice the strong difference between the empirical copulas and the Gaussian ones which are still 
widely used in financial engineering due to their convenience.
Notice also the difference between asset classes: 
Though estimated correlations are $\rho = 0.34$ for the leftmost copulas, they have much dissimilar peculiarities.

\subsubsection{Stocks} 

Centroids' main feature: 
More mass in the bottom-left corner, i.e. lower tail dependence. Stock prices tend to plummet together.

\includegraphics[width=0.233\linewidth]{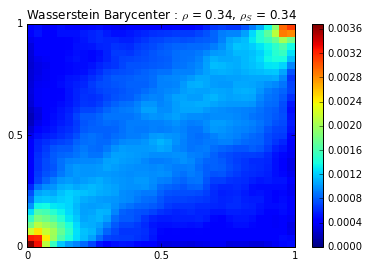}
\includegraphics[width=0.233\linewidth]{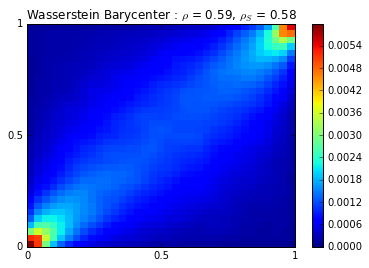}
\includegraphics[width=0.233\linewidth]{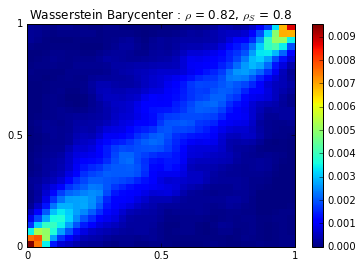}
\includegraphics[width=0.233\linewidth]{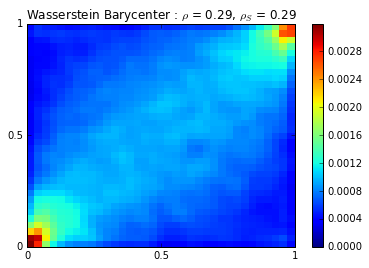}

\includegraphics[width=0.233\linewidth]{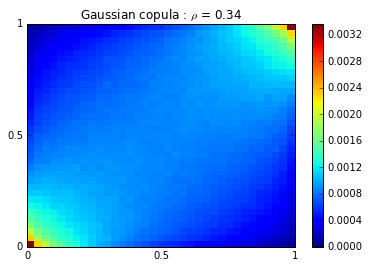}
\includegraphics[width=0.233\linewidth]{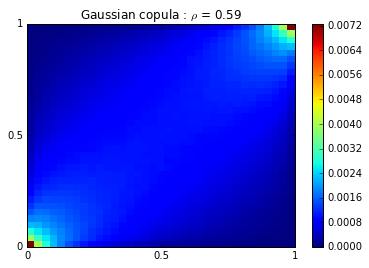}
\includegraphics[width=0.233\linewidth]{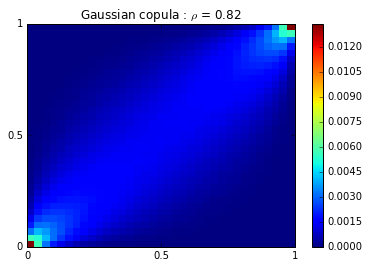}
\includegraphics[width=0.233\linewidth]{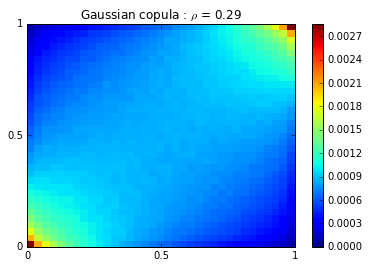}

\subsubsection{Credit default swaps}

Centroids' main feature: 
More mass in the top-right corner, i.e. upper tail dependence. Insurance cost against entities' default tends to soar in stressed market.

\includegraphics[width=0.233\linewidth]{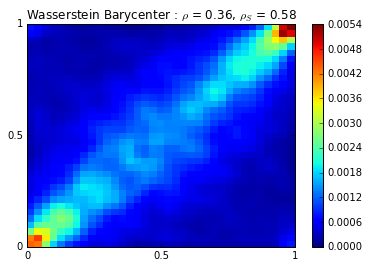}
\includegraphics[width=0.233\linewidth]{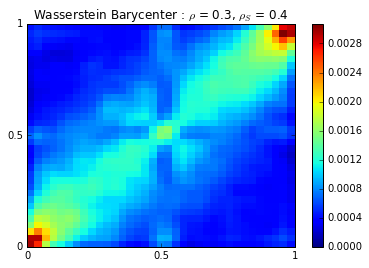}
\includegraphics[width=0.233\linewidth]{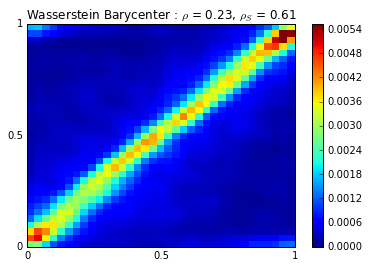}
\includegraphics[width=0.233\linewidth]{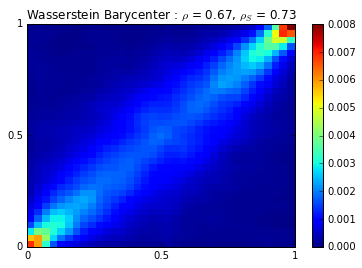}

\includegraphics[width=0.233\linewidth]{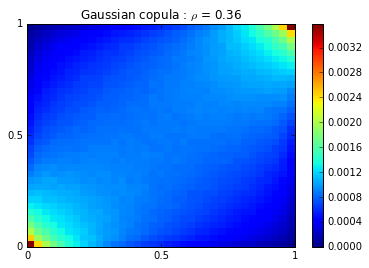}
\includegraphics[width=0.233\linewidth]{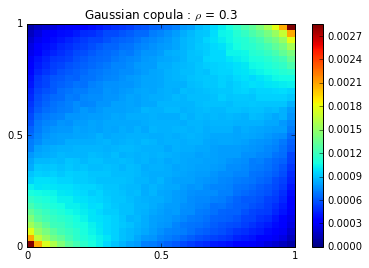}
\includegraphics[width=0.233\linewidth]{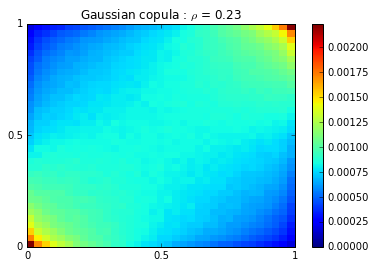}
\includegraphics[width=0.233\linewidth]{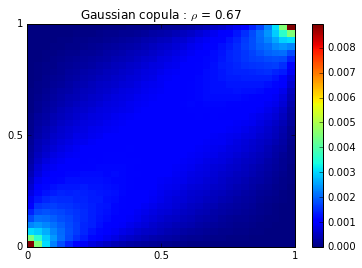}

\subsubsection{FX rates}

Centroids' main feature: 
Empirical copulas show that dependence between FX rates are various. For example, rates may exhibit either strong dependence or independence while being anti-correlated during extreme events.   


\includegraphics[width=0.233\linewidth]{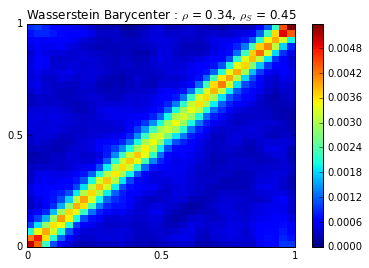}
\includegraphics[width=0.233\linewidth]{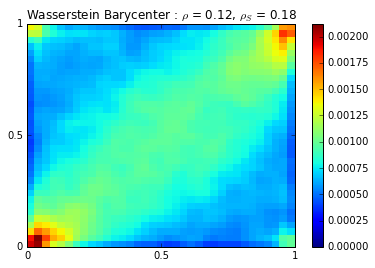}
\includegraphics[width=0.233\linewidth]{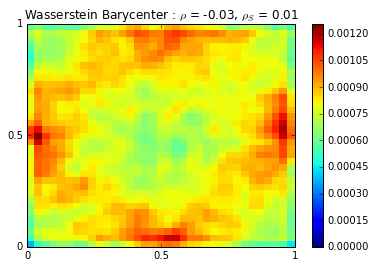}
\includegraphics[width=0.233\linewidth]{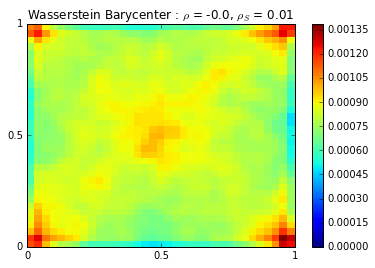}

\includegraphics[width=0.233\linewidth]{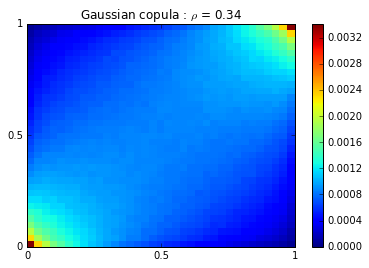}
\includegraphics[width=0.233\linewidth]{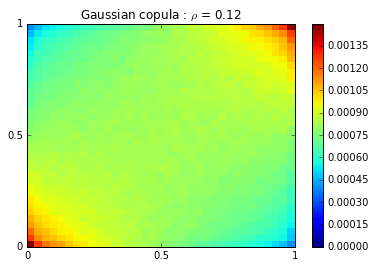}
\includegraphics[width=0.233\linewidth]{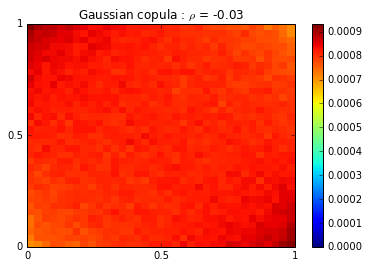}
\includegraphics[width=0.233\linewidth]{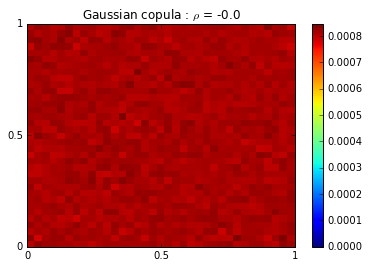}


\subsection{Answering dependence queries}

Inspired by the previous exploration results, we may want to answer such questions: (A) Which pair of assets having $\rho = 0.7$ correlation has the nearest copula to the Gaussian one? 
Though such questions can be answered by computing a likelihood for each pairs, 
our methodology stands out for dealing with non-parametric dependence patterns, and thus for questions such as: 
(B)
Which pairs of assets are both positively and negatively correlated?
(C)
Which assets occur extreme variations while those of others are relatively small, and conversely?
(D) Which pairs of assets are positively correlated for small variations but uncorrelated otherwise?

\begin{figure}
\includegraphics[width=0.24\linewidth]{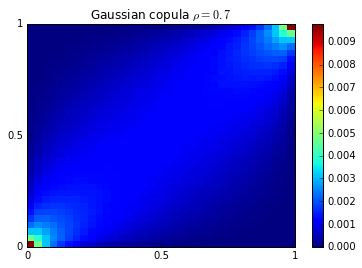}
\includegraphics[width=0.24\linewidth]{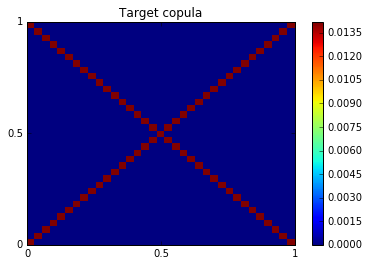}
\includegraphics[width=0.24\linewidth]{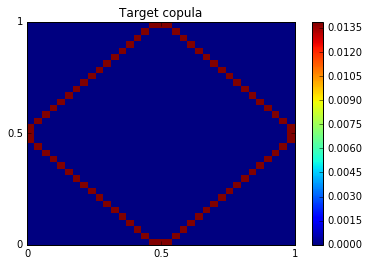}
\includegraphics[width=0.24\linewidth]{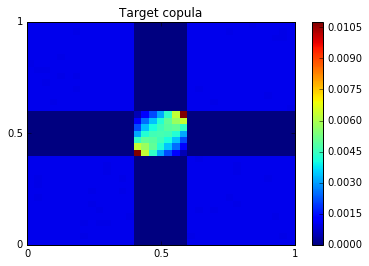}
\includegraphics[width=0.24\linewidth]{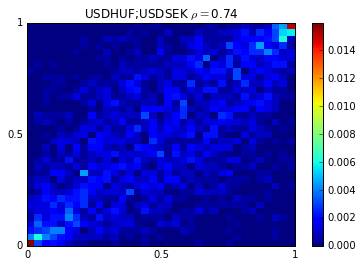}
\includegraphics[width=0.24\linewidth]{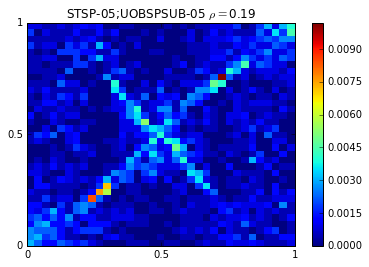}
\includegraphics[width=0.24\linewidth]{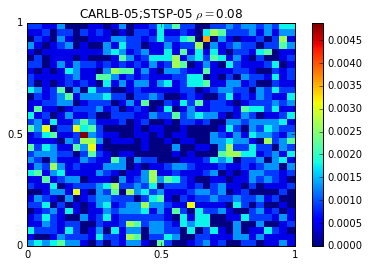}
\includegraphics[width=0.24\linewidth]{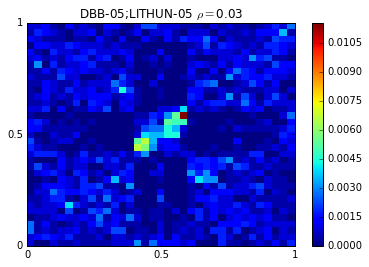}
\caption{Target copulas (simulated or handcrafted) and their respective nearest copulas which answer questions A,B,C,D}
\label{fig:gausscop}
\end{figure}

Considering a cross-asset dataset which comprises the SBF 120 components (index including the CAC 40 and 80 other highly capitalized French entities), the 500 most liquid CDS worldwide, and 80 FX rates, 
we display in Figure~\ref{fig:gausscop} the empirical copulas (alongside their respective targets) which best answer questions A,B,C,D.




\subsection{Power of TFDC}

\begin{figure}
\begin{center}
\includegraphics[width=\linewidth]{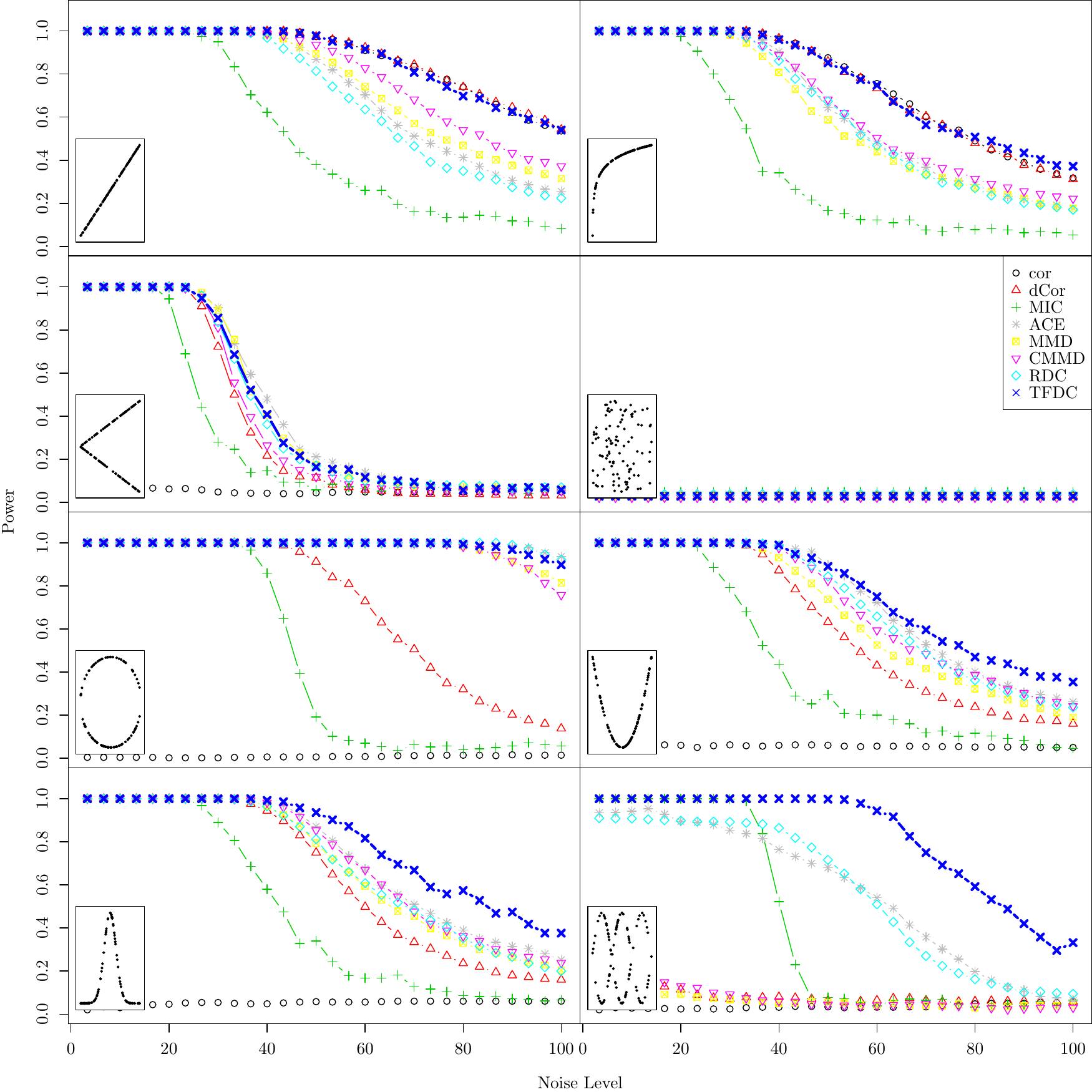}
\end{center}
\caption{Power of several dependence coefficients as a function of the noise level in eight different scenarios.
Insets show the noise-free form of each association pattern.
The coefficient power was estimated via 500 simulations with sample size 500 each}
\label{fig:power}
\end{figure}

In this experiment, we compare the empirical power of TFDC to well-known dependence coefficients such as Pearson linear correlation (cor), distance correlation (dCor) \cite{szekely2009brownian}, maximal information coefficient (MIC) \cite{reshef2011detecting}, alternating conditional expectations (ACE) \cite{breiman1985estimating}, 
maximum mean discrepancy (MMD) \cite{gretton2012kernel},
copula maximum mean discrepancy (CMMD) \cite{ghahramani2012copula},
randomized dependence coefficient (RDC) \cite{lopez2013randomized}.
Statistical power of a binary hypothesis test is the probability that the test correctly rejects the null hypothesis (H0) when the alternative hypothesis (H1) is true. In the case of dependence coefficients, we consider (H0): $X$ and $Y$ are independent; (H1): $X$ and $Y$ are dependent.
Following the numerical experiment described in \cite{simon2014comment,lopez2013randomized}, we estimate the power of the aforementioned dependence measures with simulated pairs of variables with different relationships (considered in \cite{reshef2011detecting,simon2014comment,lopez2013randomized}), but with varying levels of noise added. 
By design, TFDC aims at detecting the simulated dependence relationships. 
Thus, this dependence measure is expected to have a much higher power than coefficients such as MIC
since, according to Simon and Tibshirani in \cite{simon2014comment}, coefficients 
``which strive to have high power against all alternatives
can have low power in many important situations."
TFDC only targets the specific important situations.
Results are displayed in Figure~\ref{fig:power}.

\section{Discussion}

It is known by risk managers how dangerous it can be to rely solely on 
a correlation coefficient to measure dependence.
That is why we have proposed a novel approach to
explore, summarize and measure the pairwise correlations 
which exist between variables in a dataset.
We have also pointed out through the UCI-datasets example that 
non-trivial dependence patterns can be easily found between the 
features variables. Using these patterns as \textit{targets}
when performing correlation-based feature selection may improve results.
This idea still needs to be empirically verified.
The experiments show the benefits of the proposed method:
It allows to highlight the various dependence patterns that can be found between financial time series,
which strongly depart from the Gaussian copula widely used in financial engineering.
Though \textit{answering dependence queries} as briefly outlined is still an art, 
we plan to develop a rich language so that a user can formulate complex questions about dependence,
which will be automatically translated into copulas in order to let the methodology provide these questions accurate answers.

\begin{quote}
\begin{small}
\bibliographystyle{aaai}
\bibliography{biblio}
\end{small}
\end{quote}

\end{document}